\pdfoutput=1
\documentclass[10pt,twocolumn,letterpaper]{article}

\usepackage[pagenumbers]{cvpr}

\ifdefined\pdfsuppressptexinfo\pdfsuppressptexinfo=-1\relax\fi

\usepackage{cuted}
\usepackage{currfile}
\usepackage{fix-cm}
\usepackage{graphicx}
\usepackage{booktabs}
\usepackage{adjustbox}
\usepackage{dcolumn}
\usepackage{xspace}
\usepackage{multirow}
\usepackage{makecell}
\usepackage{tikz}
\usepackage{pgfplots}
\pgfplotsset{compat=1.18}
\usetikzlibrary{arrows.meta,backgrounds,calc,fit, positioning, shapes.geometric, decorations.pathmorphing, decorations.pathreplacing, external, spy, shapes.symbols}
\usepackage{algorithm}
\usepackage{algpseudocode}
\usepackage{xcolor}
\usepackage{colortbl}
\usepackage{pifont}
\usepackage[normalem]{ulem}
\usepackage{enumitem}
\usepackage{wrapfig}
\usepackage{listings}
\usepackage{multibib}
\makeatletter
\ifdefined\logtheLineNumber
  \patchcmd{\logtheLineNumber}{\@auxout}{\@mainaux}{}{\PackageWarning{preamble}{lineno patch failed}}
\fi
\let\mb@savedLN\@LN \let\@LN\@gobbletwo
\newcites{S}{References}
\let\@LN\mb@savedLN
\makeatother

\usepackage{amsmath,amsfonts,bm}

\def\eqref#1{equation~\ref{#1}}

\def\1{\bm{1}}

\def\vzero{{\bm{0}}}

\def\vmu{{\bm{\mu}}}
\def\vtheta{{\bm{\theta}}}

\def\vc{{\bm{c}}}

\def\ve{{\bm{e}}}

\def\vq{{\bm{q}}}

\def\vv{{\bm{v}}}

\def\vz{{\bm{z}}}

\def\mI{{\bm{I}}}

\def\mZ{{\bm{Z}}}

\DeclareMathAlphabet{\mathsfit}{\encodingdefault}{\sfdefault}{m}{sl}
\SetMathAlphabet{\mathsfit}{bold}{\encodingdefault}{\sfdefault}{bx}{n}

\def\gD{{\mathcal{D}}}

\def\gI{{\mathcal{I}}}

\newcommand{\E}{\mathbb{E}}

\newcommand{\R}{\mathbb{R}}

\definecolor{cvprblue}{rgb}{0.21,0.49,0.74}
\usepackage[pagebackref,breaklinks,colorlinks,allcolors=cvprblue]{hyperref}

\AtEndPreamble{%
  \crefname{proposition}{Prop.}{Props.}%
  \Crefname{proposition}{Prop.}{Props.}%
  \crefname{algorithm}{Alg.}{Algs.}%
  \Crefname{algorithm}{Alg.}{Algs.}%
  \crefname{figure}{Fig.}{Figs.}%
  \Crefname{figure}{Fig.}{Figs.}%
  \crefname{subfigure}{Fig.}{Figs.}%
  \Crefname{subfigure}{Fig.}{Figs.}%
  \crefname{table}{Tab.}{Tabs.}%
  \Crefname{table}{Tab.}{Tabs.}%
  \crefname{section}{Sec.}{Secs.}%
  \Crefname{section}{Sec.}{Secs.}%
  \crefname{subsection}{Sec.}{Secs.}%
  \Crefname{subsection}{Sec.}{Secs.}%
  \crefname{subsubsection}{Sec.}{Secs.}%
  \Crefname{subsubsection}{Sec.}{Secs.}%
  \crefname{equation}{Eq.}{Eqs.}%
  \Crefname{equation}{Eq.}{Eqs.}%
  \crefname{appendix}{App.}{Apps.}%
  \Crefname{appendix}{App.}{Apps.}%
}

\DeclareMathSizes{5}{5}{3.5}{3}%
\newcommand{\method}{RSD\xspace}%

\newcommand{\ignorethis}[1]{}

\newcommand{\Reals      }     {{\textrm{I\kern-0.18em R}}}

\newcommand{\change     } [1] {\mbox{{\footnotesize $\Delta$} \kern-3pt}#1}

\newlength{\x}

\definecolor{darkred}{rgb}{0.7,0.1,0.1}
\definecolor{darkgreen}{rgb}{0.1,0.6,0.1}
\definecolor{cyan}{rgb}{0.7,0.0,0.7}
\definecolor{otherblue}{rgb}{0.1,0.4,0.8}
\definecolor{maroon}{rgb}{0.76,.13,.28}
\definecolor{burntorange}{rgb}{0.81,.33,0}
\definecolor{cyan_real}{rgb}{0.0,1.0,1.0}
\definecolor{purple_real}{rgb}{0.5,0.0,0.5}

\newcommand{\N}{\mathcal{N}}

\newcommand{\probP}{\text{I\kern-0.15em P}}

\newcommand{\gridprompt}[1]{{\usefont{T1}{ppl}{m}{it}\strut
  {\usefont{T1}{ptm}{m}{n}\textquotedblleft}#1{\usefont{T1}{ptm}{m}{n}\textquotedblright}}}

\providecommand{\zoominsetfrac}{0.42}%
\providecommand{\zoomcolor}{zoomcolor@default}%
\definecolor{zoomcolor@default}{HTML}{F2A900}%
\providecommand{\zoommarks}{}%
\definecolor{markcolor}{HTML}{D81B60}%
\tikzset{defectmark/.style={markcolor,line width=0.6pt}}%
\newcommand{\marktile}[2]{\ifx\zoommarks\empty\includegraphics[width=#1]{#2}\else
  \begin{tikzpicture}
    \node[anchor=south west,inner sep=0] (img) at (0,0) {\includegraphics[width=#1]{#2}};
    \begin{scope}[x={(img.south east)},y={(img.north west)}]\zoommarks\end{scope}
  \end{tikzpicture}\fi}%
\makeatletter
\newlength{\zt@tilew}
\newlength{\zt@roih}
\newlength{\zt@insetw}
\newlength{\zt@inseth}
\newlength{\zt@bigw}
\newlength{\zt@margin}

\newlength{\zoompopx}
\newlength{\zoompopy}
\newlength{\zt@mx}
\newlength{\zt@my}
\newif\ifzt@pop

\newcommand{\zoomtile}[7]{%
  \edef\zt@tilewd{\the\dimexpr#1\relax}\setlength{\zt@tilew}{\zt@tilewd}%
  \begin{tikzpicture}
    \node[anchor=south west,inner sep=0] (img) at (0,0) {\includegraphics[width=\zt@tilew]{#2}};
    \ifdim\dimexpr\zoompopx+\zoompopy\relax>0pt
      \zt@poptrue \setlength{\zt@mx}{-\zoompopx}\setlength{\zt@my}{-\zoompopy}%
      \useasboundingbox (img.south west) rectangle (img.north east);
    \else
      \zt@popfalse \setlength{\zt@mx}{\zt@margin}\setlength{\zt@my}{\zt@margin}%
    \fi
    \begin{scope}[x={(img.south east)},y={(img.north west)}]
      \coordinate (roi-nw) at (#3,{1-(#4)});
      \coordinate (roi-ne) at ({(#3)+(#5)},{1-(#4)});
      \coordinate (roi-sw) at (#3,{1-(#4)-(#6)});
      \coordinate (roi-se) at ({(#3)+(#5)},{1-(#4)-(#6)});
    \end{scope}
    \pgfpointdiff{\pgfpointanchor{roi-sw}{center}}{\pgfpointanchor{roi-nw}{center}}%
    \edef\zt@roihd{\the\pgf@y}\setlength{\zt@roih}{\zt@roihd}%
    \pgfmathsetmacro{\ztmag}{\zoominsetfrac/(#5)}%
    \setlength{\zt@insetw}{\zoominsetfrac\zt@tilew}%
    \setlength{\zt@inseth}{\ztmag\zt@roih}%
    \setlength{\zt@bigw}{\ztmag\zt@tilew}%
    \edef\zt@corner{#7}%
    \def\zt@NW{north west}\def\zt@NE{north east}%
    \def\zt@SW{south west}\def\zt@SE{south east}%
    \ifx\zt@corner\zt@NW
      \coordinate (cw-nw) at ($(img.north west)+(\zt@mx,-\zt@my)$);
      \coordinate (cw-se) at ($(cw-nw)+(\zt@insetw,-\zt@inseth)$);
      \def\zt@roinear{roi-ne}\def\zt@cwnear{cw-se}%
    \else\ifx\zt@corner\zt@NE
      \coordinate (cw-ne) at ($(img.north east)+(-\zt@mx,-\zt@my)$);
      \coordinate (cw-nw) at ($(cw-ne)+(-\zt@insetw,0pt)$);
      \coordinate (cw-se) at ($(cw-nw)+(\zt@insetw,-\zt@inseth)$);
      \def\zt@roinear{roi-nw}\def\zt@cwnear{cw-sw}%
    \else\ifx\zt@corner\zt@SW
      \coordinate (cw-sw) at ($(img.south west)+(\zt@mx,\zt@my)$);
      \coordinate (cw-nw) at ($(cw-sw)+(0pt,\zt@inseth)$);
      \coordinate (cw-se) at ($(cw-nw)+(\zt@insetw,-\zt@inseth)$);
      \def\zt@roinear{roi-se}\def\zt@cwnear{cw-ne}%
    \else
      \coordinate (cw-se) at ($(img.south east)+(-\zt@mx,\zt@my)$);
      \coordinate (cw-nw) at ($(cw-se)+(-\zt@insetw,\zt@inseth)$);
      \def\zt@roinear{roi-sw}\def\zt@cwnear{cw-nw}%
    \fi\fi\fi%
    \coordinate (cw-sw) at (cw-nw |- cw-se);
    \coordinate (cw-ne) at (cw-se |- cw-nw);
    \coordinate (roi-near) at (\zt@roinear);
    \coordinate (cw-near) at (\zt@cwnear);
    \coordinate (zt-vec) at ($(roi-nw)-(img.south west)$);
    \coordinate (zt-p2) at ($(cw-nw)-\ztmag*(zt-vec)$);
    \ifzt@pop\fill[white] (cw-nw) rectangle (cw-se);\fi
    \begin{scope}
      \clip (cw-nw) rectangle (cw-se);
      \node[anchor=south west,inner sep=0] at (zt-p2) {\includegraphics[width=\zt@bigw]{#2}};
    \end{scope}
    \draw[\zoomcolor,line width=0.5pt,opacity=0.7] (roi-near) -- (cw-near);
    \draw[white,line width=1.2pt] (roi-nw) rectangle (roi-se);
    \draw[\zoomcolor,line width=0.6pt] (roi-nw) rectangle (roi-se);
    \ifzt@pop
      \draw[white,line width=1.6pt] (cw-nw) rectangle (cw-se);
      \draw[\zoomcolor,line width=0.9pt] (cw-nw) rectangle (cw-se);
    \else
      \draw[white,line width=1.2pt] (cw-nw) rectangle (cw-se);
      \draw[\zoomcolor,line width=0.6pt] (cw-nw) rectangle (cw-se);
    \fi
    \begin{scope}[x={(img.south east)},y={(img.north west)}]\zoommarks\end{scope}
  \end{tikzpicture}%
}
\makeatother

\algnewcommand{\IfThen}[2]{\State \algorithmicif\ #1\ \algorithmicthen\ #2}

\newcommand{\veps}{{\bm{\epsilon}}}

\tikzexternaldisable
\newcommand{\figcachebegin}[1]{\tikzsetfigurename{#1-}\tikzexternalenable}
\newcommand{\figcacheend}{\tikzexternaldisable}

\title{Retrieval-Seeded Diffusion for Semantic Flow Models}

\author{Omer Seri \qquad Shimon Malnick \qquad Shai Avidan\\[0.5em]
Tel Aviv University\\[0.3em]
{\small\url{https://omerseri.github.io/RSD}}\\[-10pt]
}

\begin{document}
\maketitle
\begin{strip}
  \centering
  \newcommand{\ttfont}{\sffamily}%
  \newcommand{\ttbold}{\sffamily\bfseries}%
  \definecolor{ttbase}{HTML}{EB6834}%
  \definecolor{ttours}{HTML}{2A78D6}%
  \definecolor{ttmu}{HTML}{B0003A}%
  \colorlet{ttgt}{darkgreen!85!black}%
  \newcommand{\ttdir}{figures/teaser_traj/img}%
  \newcommand{\ttx}[3]{\draw[#2, line width=1.3pt, line cap=round]
    ($#1+(-#3,-#3)$) -- ($#1+(#3,#3)$) ($#1+(-#3,#3)$) -- ($#1+(#3,-#3)$);}%
  \expandafter\def\csname teaserqual@base\endcsname{50}%
\expandafter\def\csname teaserqual@ours\endcsname{8}%
\expandafter\def\csname teaserqual@ratio\endcsname{6.25}%
\expandafter\def\csname teaserqual@rae@p03797\endcsname{A red fire hydrant sitting on the side of a road.}%
\expandafter\def\csname teaserqual@rae@p02189\endcsname{A cross country skier on the slopes on a sunny day}%
\expandafter\def\csname teaserqual@rae@p01959\endcsname{A train traveling through the countryside next to a dirt road.}%
\expandafter\def\csname teaserqual@rae@p02377\endcsname{A person riding a brown horse with blonde hair in a green field.}%
\expandafter\def\csname teaserqual@rae@p02854\endcsname{A red and white boat floating along a river.}%
\expandafter\def\csname teaserqual@rae@p03087\endcsname{A large pizza sitting on a pizza pan with a slice missing.}%
\expandafter\def\csname teaserqual@sefiart@p00863\endcsname{A young girl with a blue bow in her hair wears a yellow hat and looks at the viewer.}%
\expandafter\def\csname teaserqual@sefiart@p01937\endcsname{A painting of a bird and a lion with a dark background.}%
\expandafter\def\csname teaserqual@sefico@p00158\endcsname{An outdoor area with many animals including a zebra and a rhinoceros.}%
\expandafter\def\csname teaserqual@svgkco@p01675\endcsname{A half eaten banana being held by a human hand.}%
  \newcommand{\ttv}[1]{\csname teaserqual@#1\endcsname}%
  \newcommand{\ttgridw}{0.57}%
  \newcommand{\ttpanelw}{0.265}%
  \newcommand{\ttlabw}{0.17}%
  \newcommand{\ttcolsep}{1.5pt}%
  \newcommand{\ttpromptsep}{2.5pt}%
  \newcommand{\ttrowsep}{3pt}%
  \newcommand{\ttpromptfont}{\scriptsize}%
  \newcommand{\ttframe}{0.3pt}%
  \newcommand{\ttlabfont}{\footnotesize}%
  \newcommand{\ttbadgefont}{\scriptsize\bfseries}%
  \newcommand{\ttbadgesep}{3pt}%
  \ifdefined\figcachebegin
    \tikzset{external/optimize=false}%
    \pgfkeys{/pgf/images/include external/.code={\xdef\pgfexternaldepth{\the\dimexpr\pgfexternaldepth-2pt\relax}\includegraphics[trim=0 2pt 0 2pt]{#1}}}%
    \figcachebegin{teaser_traj}%
  \fi
  \newsavebox{\ttgridbox}%
  \sbox{\ttgridbox}{\ifdefined\figcachebegin\tikzset{external/export=false}\fi%
    \begin{minipage}[c]{\ttgridw\linewidth}
      \setlength{\tabcolsep}{\ttcolsep}%
      \renewcommand{\arraystretch}{0}%
      \setlength{\fboxsep}{0pt}\setlength{\fboxrule}{\ttframe}%
      \newcommand{\ttimg}[2]{\fbox{\includegraphics[width=\dimexpr\ttpanelw\linewidth-2\fboxrule\relax]{figures/teaser_qual/#1/#2.jpg}}}%
      \newcommand{\ttprompt}[1]{\parbox[b]{\ttpanelw\linewidth}{\centering\ttpromptfont\gridprompt{\ttv{#1}}}}%
      \newcommand{\ttlab}[2]{\parbox[b][\ttpanelw\linewidth][c]{\ttlabw\linewidth}{\centering\color{#1}#2}}%
      \newcommand{\ttbadge}{\tikz[baseline=(t.base)]\node[fill=otherblue, text=white, rounded corners=2pt,
        inner xsep=2.5pt, inner ysep=1.6pt, font=\ttbadgefont] (t) {\ttv{ratio}$\times$ faster};}%
      \newcommand{\ttrow}[1]{%
        \ttimg{sefi_coco/p00158}{#1} & \ttimg{sefi_art/p00863}{#1} & \ttimg{svg1k_coco/p01675}{#1} &}%
      \begin{tabular}{@{}cccc@{}}%
        \ttprompt{sefico@p00158} & \ttprompt{sefiart@p00863} & \ttprompt{svgkco@p01675} & \\[\ttpromptsep]
        \ttrow{base} \ttlab{ttbase}{\ttlabfont Baseline} \\[\ttrowsep]%
        \ttrow{ours} \ttlab{otherblue}{\ttlabfont\textbf{\method{} (ours)}\\[\ttbadgesep]\ttbadge} \\
      \end{tabular}
    \end{minipage}
  }%
  \pgfmathsetmacro{\ttyscale}{(\ht\ttgridbox+\dp\ttgridbox)/4.64cm}%
  \newcommand{\ttgap}{6pt}%
  \pgfmathsetmacro{\ttscale}{((1-\ttgridw)*\linewidth-\ttgap)/(6.17*1cm)}%
  \newcommand{\ttnbside}{1.0cm}%
  \newcommand{\ttnbframe}{1.3pt}%
  \newcommand{\tthlframe}{2.6pt}%
  \newcommand{\ttmusize}{1.6pt}%
  \newcommand{\ttmufont}{\small}%
  \newcommand{\ttlegfont}{\tiny}%
  \newcommand{\ttsteps}[6]{\foreach \i in {0,...,#2} {
      \pgfmathsetmacro{\ttt}{\i/#2}
      \coordinate (ttqa) at ($(#3)!\ttt!(#4)$); \coordinate (ttqb) at ($(#4)!\ttt!(#5)$);
      \coordinate (ttqc) at ($(#5)!\ttt!(#6)$);
      \coordinate (ttra) at ($(ttqa)!\ttt!(ttqb)$); \coordinate (ttrb) at ($(ttqb)!\ttt!(ttqc)$);
      \coordinate (#1\i) at ($(ttra)!\ttt!(ttrb)$);}}%
  \newcommand{\tttaper}[5]{%
    \pgfmathtruncatemacro{\ttnm}{#2-1}\pgfmathtruncatemacro{\ttnmm}{#2-2}
    \foreach \i in {1,...,\ttnm} {
      \pgfmathtruncatemacro{\j}{\i-1}
      \pgfmathsetmacro{\ttw}{#4+(#5-#4)*\i/#2}
      \draw[#3, line width=\ttw pt, line cap=round] (#1\j) -- (#1\i);}
    \draw[#3, line width=#5pt, -{Latex[round, length=8pt, width=7.5pt]}, shorten >=2.4pt] (#1\ttnm) -- (#1#2);
    \fill[#3] (#1#2) circle (2.7pt);
    \foreach \i in {1,...,\ttnmm} \fill[white, opacity=0.85] (#1\i) circle (0.75pt);}%
  \newcommand{\tttile}[4]{%
    \node[inner sep=0pt, draw=#4, line width=\ttnbframe, rounded corners=1.5pt] at #1 {\includegraphics[width=#2]{#3}};}%
  \edef\ttaxis{\the\fontdimen22\textfont2}%
  \begin{tikzpicture}[baseline={([yshift=-\ttaxis]current bounding box.center)},%
      xscale=\ttscale, yscale=\ttyscale,
      font=\ttfont\footnotesize,
      nb/.style={circle, fill=ttmu, draw=white, line width=0.5pt, inner sep=1.4pt},
    ]
    \tikzset{declare function={
      ttsrc(\y) = 0.62*exp(-((\y-2.10)^2)/0.605);
      tttgt(\y) = 0.40*exp(-((\y-0.75)^2)/0.0512) + 0.55*exp(-((\y-1.45)^2)/0.0722)
                + 0.85*exp(-((\y-2.78)^2)/0.12) + 0.38*exp(-((\y-3.60)^2)/0.03);}}%
    \def\ttxs{0.25}\def\ttxt{4.00}\def\ttylo{0.38}\def\ttyhi{3.88}
    \fill[black!2.5, draw=black!10, line width=0.5pt, rounded corners=6pt] (-0.12,-0.12) rectangle (6.05,4.52);
    \foreach \ya/\yb in {0.80/0.62, 1.05/0.80, 1.30/1.12, 1.52/1.38, 1.72/1.55, 1.92/1.78,
                         2.62/2.42, 2.82/2.70, 3.02/3.02, 3.25/3.28, 3.48/3.55, 3.70/3.70} {
      \pgfmathsetmacro{\ttfa}{\ttxs+ttsrc(\ya)}\pgfmathsetmacro{\ttfb}{\ttxt+tttgt(\yb)}
      \draw[black!13, line width=0.6pt, -{Latex[length=3pt]}, shorten >=1pt]
        (\ttfa,\ya) .. controls +(1.3,0) and +(-1.3,0) .. (\ttfb,\yb);}
    \shade[left color=black!22, right color=black!4] (\ttxs,\ttylo) -- plot[domain=\ttylo:\ttyhi, samples=90, smooth, variable=\y]
      ({\ttxs+ttsrc(\y)}, \y) -- (\ttxs,\ttyhi) -- cycle;
    \draw[black!50, line width=0.9pt] plot[domain=\ttylo:\ttyhi, samples=90, smooth, variable=\y] ({\ttxs+ttsrc(\y)}, \y);
    \draw[black!35, line width=0.6pt] (\ttxs,\ttylo) -- (\ttxs,\ttyhi);
    \shade[left color=ttgt!30, right color=ttgt!6] (\ttxt,\ttylo) -- plot[domain=\ttylo:\ttyhi, samples=160, smooth, variable=\y]
      ({\ttxt+tttgt(\y)}, \y) -- (\ttxt,\ttyhi) -- cycle;
    \begin{scope}
      \clip (\ttxt,2.25) rectangle (\ttxt+1,3.15);%
      \shade[left color=ttgt!48, right color=ttgt!18] (\ttxt,\ttylo) -- plot[domain=\ttylo:\ttyhi, samples=160, smooth, variable=\y]
        ({\ttxt+tttgt(\y)}, \y) -- (\ttxt,\ttyhi) -- cycle;
    \end{scope}
    \draw[ttgt!75!black, line width=0.9pt] plot[domain=\ttylo:\ttyhi, samples=160, smooth, variable=\y] ({\ttxt+tttgt(\y)}, \y);
    \draw[black!35, line width=0.6pt] (\ttxt,\ttylo) -- (\ttxt,\ttyhi);
    \node[inner sep=1pt, font=\ttbold\footnotesize, text=black!70, anchor=base] at (1.12,4.08) {Source $\mathcal{N}(\mathbf{0},\mathbf{I})$};
    \node[inner sep=1pt, font=\ttbold\footnotesize, text=black!70, anchor=base] at (4.30,4.08) {Target};
    \node[inner sep=1pt, font=\ttfont\scriptsize, text=black!50, anchor=north] at (\ttxs,\ttylo-0.04) {$t = 0$};
    \node[inner sep=1pt, font=\ttfont\scriptsize, text=black!50, anchor=north] at (\ttxt,\ttylo-0.04) {$t = 1$};
    \pgfmathsetmacro{\ttpx}{\ttxs+ttsrc(2.40)}\coordinate (eps) at (\ttpx,2.40);%
    \pgfmathsetmacro{\ttpx}{\ttxt+tttgt(2.88)}\coordinate (rsd) at (\ttpx,2.88);%
    \pgfmathsetmacro{\ttpx}{\ttxt+tttgt(3.35)}\coordinate (base) at (\ttpx,3.35);%
    \pgfmathsetmacro{\ttpx}{\ttxt+(tttgt(2.40)+tttgt(2.62)+tttgt(3.05))/3}\coordinate (mu) at (\ttpx,2.69);
    \pgfmathsetmacro{\ttux}{1/(\ttscale*1cm)}\pgfmathsetmacro{\ttuy}{4.64/(\ht\ttgridbox+\dp\ttgridbox)}%
    \pgfmathsetmacro{\ttgx}{-0.12+(1-\ttgridw)*\linewidth*\ttux}%
    \pgfmathsetmacro{\ttpw}{\ttpanelw*\ttgridw*\linewidth}%
    \pgfmathsetmacro{\ttgxr}{\ttgx+\ttpw*\ttux}%
    \pgfmathsetmacro{\ttyra}{-0.12}\pgfmathsetmacro{\ttyrb}{-0.12+\ttpw*\ttuy}%
    \pgfmathsetmacro{\ttyba}{-0.12+(\ttpw+\ttrowsep)*\ttuy}\pgfmathsetmacro{\ttybb}{\ttyba+\ttpw*\ttuy}%
    \pgfmathsetmacro{\ttnbdy}{(\ttnbside+3pt)*\ttuy}%
    \pgfmathsetmacro{\ttnbhx}{0.5*\ttnbside*\ttux}%
    \pgfmathsetmacro{\ttnbx}{5.97-\ttnbhx}%
    \foreach \y/\k/\f in {3.05/1/c158_nb1, 2.62/0/c158_nb2, 2.40/-1/c158_nb3} {%
      \pgfmathsetmacro{\ttnx}{\ttxt+tttgt(\y)}
      \pgfmathsetmacro{\ty}{2.83+\k*\ttnbdy}
      \draw[ttmu!50, line width=0.5pt] (\ttnx,\y) -- (mu);
      \draw[ttmu!45, line width=0.5pt] (\ttnx,\y) -- (\ttnbx-\ttnbhx,\ty);
      \tttile{(\ttnbx,\ty)}{\ttnbside}{\ttdir/\f.jpg}{ttmu!80}
      \node[nb] at (\ttnx,\y) {};}
    \pgfmathsetmacro{\ttnbtop}{2.83+\ttnbdy+0.5*\ttnbside*\ttuy+0.02}%
    \node[font=\ttbold\scriptsize, text=ttmu, anchor=south] at (\ttnbx,\ttnbtop) {Retrieved};
    \draw[fill=white, draw=black!15, line width=0.5pt, rounded corners=3pt] (4.86,0.30) rectangle (6.0,1.22);
    \foreach \c/\n/\s/\y in {ttbase/Baseline/\ttv{base}/0.97, ttours/\method/\ttv{ours}/0.56} {
      \draw[\c, line width=1.8pt, -{Latex[round, length=4.5pt, width=4.5pt]}] (4.94,\y) -- (5.20,\y);
      \node[anchor=west, align=left, inner sep=0.5pt, font=\ttfont\ttlegfont, text=\c!90!black]
        at (5.22,\y) {\textbf{\n}\\[-0.6pt]\s{} steps};}
    \coordinate (z0) at ($(eps)!0.2!(mu)$);%
    \coordinate (ttlo) at (0,\ttylo);
    \draw[ttours!65, line width=0.85pt, dash pattern=on 1.2pt off 1.4pt] (z0 |- ttlo) -- (z0);
    \node[font=\ttfont\scriptsize, text=ttours!90!black, anchor=north] at ($(z0 |- ttlo)+(0,-0.04)$) {$t = t_0$};
    \coordinate (bc1) at ($(eps)+(0.45,0.95)$);  \coordinate (bc2) at ($(base)+(-1.1,0.45)$);%
    \coordinate (rc1) at ($(z0)+(0.9,0.42)$);    \coordinate (rc2) at ($(rsd)+(-0.9,0.28)$);%
    \ttsteps{bs}{24}{eps}{bc1}{bc2}{base}%
    \ttsteps{rs}{8}{z0}{rc1}{rc2}{rsd}
    \draw[ttmu!75, line width=1.1pt, -{Stealth[length=6pt, width=5pt]}, shorten >=4.5pt, shorten <=6pt] (eps) -- (mu);%
    \coordinate (da) at ($(eps)!0.15cm!-90:(mu)$); \coordinate (db) at ($(z0)!0.15cm!90:(eps)$);%
    \draw[ttmu, line width=2.2pt, -{Latex[round, length=5pt, width=6pt]}, shorten <=5pt, shorten >=4.5pt] (da) -- (db);%
    \tttaper{bs}{24}{ttbase}{0.9}{2.6}
    \tttaper{rs}{8}{ttours}{0.9}{2.6}
    \foreach \r/\o in {9.5/0.025, 7/0.04, 4.7/0.06} \fill[ttmu, opacity=\o] (mu) circle ({\r*\ttmusize/1.6});%
    \node[star, star points=5, star point ratio=2.2, fill=ttmu, draw=white, line width=0.5pt, inner sep=\ttmusize] at (mu) {};
    \node[anchor=north, inner sep=0.5pt, text=ttmu, font=\ttmufont] at ($(mu)+(0.02,-0.17)$) {$\vmu$};%
    \ttx{(eps)}{black!85}{0.08}
    \node[circle, fill=white, draw=ttours, line width=1.1pt, inner sep=1.7pt] at (z0) {};
    \node[anchor=east, inner sep=1.5pt] at ($(eps)+(-0.09,0)$) {$\veps$};
    \node[anchor=north west, inner sep=1.5pt, text=ttours] at ($(z0)+(0.04,-0.05)$) {$\vz_{t_0}$};
    \node[font=\ttbold\scriptsize, text=ttmu, anchor=north west, inner sep=1pt, align=left]
      at ($(z0)+(-0.46,-0.31)$) {shift\\[0pt]toward $\vmu$};%
    \node[anchor=south west, inner sep=0pt, outer sep=0pt] at (\ttgx,-0.12) {\usebox{\ttgridbox}};
    \draw[ttbase, line width=\tthlframe] (\ttgx,\ttyba) rectangle (\ttgxr,\ttybb);
    \draw[ttours, line width=\tthlframe] (\ttgx,\ttyra) rectangle (\ttgxr,\ttyrb);
    \pgfmathsetmacro{\ttgxe}{\ttgx+\ttgridw*\linewidth*\ttux}%
    \pgfresetboundingbox\path[use as bounding box] (-0.12,-0.12) rectangle (\ttgxe,4.52);
    \ifdefined\figcachebegin\global\advance\csname pgf@picminy\endcsname-2pt\global\advance\csname pgf@picmaxy\endcsname2pt\fi%
  \end{tikzpicture}%
  \ifdefined\figcacheend\figcacheend\fi
  \captionsetup{hypcap=false}%
  
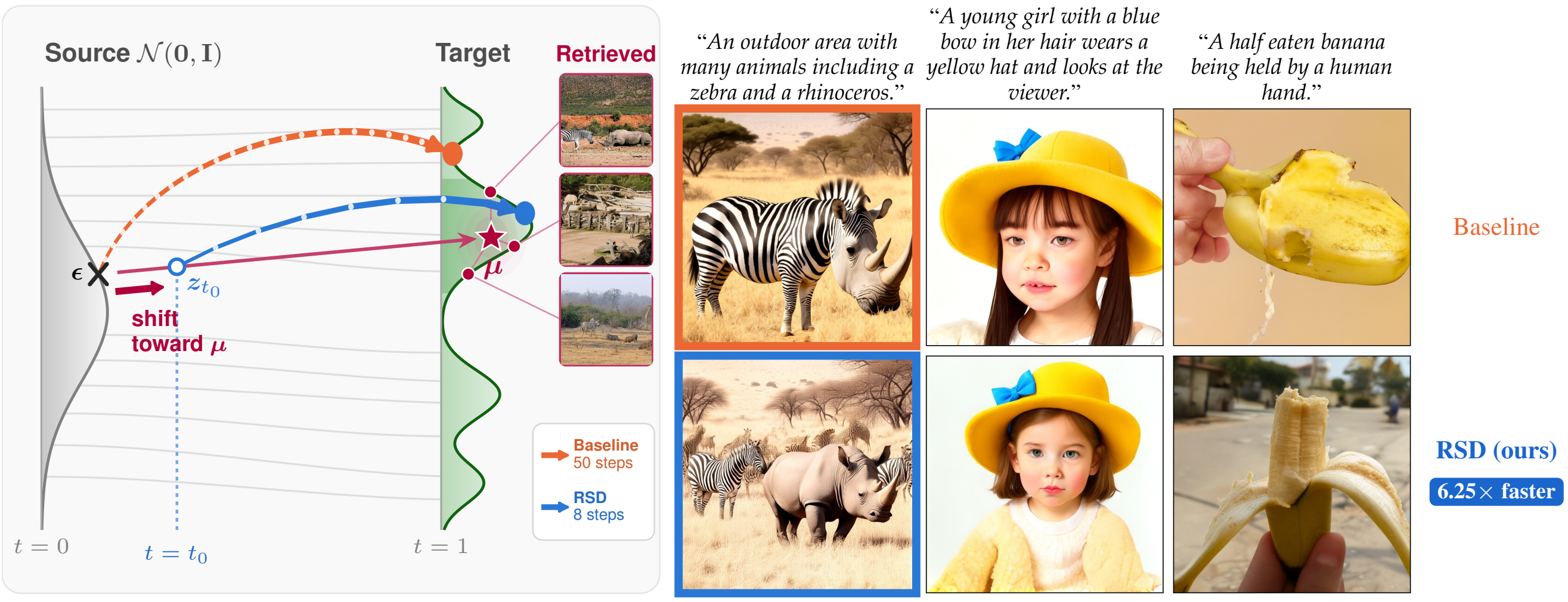
\captionof{figure}{\textbf{\method shifts the seed $\veps$ toward an anchor.} The anchor $\vmu$ is a
    weighted mean of the semantic latents of images retrieved for the prompt. \method moves $\veps$ a
    short step $t_0$ towards it and lets the frozen flow model run from there. Left: from the same $\veps$,
    the released sampler (orange) might land off the prompt's mode, while \method (blue) reaches it in fewer
    steps, drawn as segments. Right: the released sampler (SeFi-Image) with \ttv{base} steps (top) and \method with
    \ttv{ours} (bottom).
    }%
  \captionsetup{hypcap=true}%
  \label{fig:teaser_traj}
\end{strip}
\begin{abstract}
Diffusion models are trained to map a source distribution (usually, a Gaussian) to a target distribution (say, of natural images).
The Gaussian distribution is chosen because it is easy to sample from, and the diffusion process is often guided by a text prompt in order to achieve a specific outcome.
A recent line of text-to-image models operates in semantic latent spaces rather than a pixel compression one, so that nearby latents share content rather than appearance, and it is these semantic models that we focus on.
Often, the images we want belong to the domain of a specific dataset, such as photographs, anime or paintings, yet sampling still starts from the same Gaussian, which takes neither the prompt nor the dataset into account.
We propose a method that generates images more similar to the dataset's domain, in fewer steps, by enhancing the Gaussian using the dataset, and call this approach Retrieval-Seeded Diffusion, or RSD.
Specifically, we store a semantic latents dataset of captioned images.
Given a new text prompt, we find its nearest neighbours in the text embedding space, use them to retrieve their semantic latent image representation counterparts, and use the mean of the semantic latent image vectors to guide the initial seed for the diffusion model.
Relying on a dataset requires no training and a new dataset can always be introduced or updated.
Experiments on three frozen semantic backbones, Scale-RAE, SVG-T2I and SeFi-Image, and three retrieved datasets, show that RSD lowers FID to the target dataset in eight of nine settings, by up to 57\%, at $6.25\times$ fewer sampling steps, with comparable prompt alignment.
\end{abstract}

\section{Introduction}

\label{sec:intro}
Diffusion and flow models now dominate text-to-image generation~\cite{lipman2023flow,esser2024scaling,wu2025qwenimage}.
Most of these methods operate in a lower-dimensional latent space,
learned by an autoencoder that is trained separately from the generative model for pixel reconstruction~\cite{rombach2021highresolution}.
A recent line of work instead generates in a semantic latent space built on a pretrained representation encoder~\cite{zheng2025rae,tong2026scalerae,shi2025svg,shi2025svgt2i},
and reports faster convergence in training.
In such a space, images with similar content get nearby latents, even when they look different.
The move to a semantic space has nevertheless left the prior untouched, and every sample still starts from $\N(\vzero, \mI)$, whatever the prompt.
Prompt-dependent priors have been
designed~\cite{issachar2025designing} or learned~\cite{kim2026bettersourcebetterflow},
but the model has to be trained with such a prior from the start, and it cannot be replaced after training.
Instead, we look for a prior that can be added
on top of a trained model,
as a knob that can be turned up, down or off.
A prompt can be compared with the captions
of a corpus of images in a text embedding space,
this tells us which images resemble it.
In a latent space trained for pixel reconstruction,
the latents of these images would mostly record their appearance and not their content.
When the model generates in a semantic latent space instead,
the latents carry what their captions describe and a proximity in captions leads to a proximity in latents.
Moreover, the images we want often belong to the domain of a specific dataset, such as photographs,
anime or paintings, and a corpus drawn from that domain describes it.
This leads us to ask:

\emph{Can we use a corpus of captioned images from a target domain to create a prompt-dependent
prior that steers a frozen semantic flow model towards that domain?}

\begin{figure}[t]
  \centering
  \input{figures/traj_curve/values}%
  \newcommand{\trdir}{figures/traj_curve/girl}%
  \newcommand{\trpromptkey}{prompt@girl}%
  \newcommand{\trw}{0.87\linewidth}%
  \newcommand{\trh}{0.66\linewidth}%
  \newcommand{\trxmax}{52}%
  \newcommand{\trymin}{15}%
  \newcommand{\trymax}{120}%
  \newcommand{\trthumb}{0.135\linewidth}%
  \newcommand{\trpad}{0.5pt}%
  \newcommand{\trframe}{1.2pt}%
  \newcommand{\trcolgap}{3pt}%
  \newcommand{\trrowgap}{3pt}%
  \newcommand{\trinset}{3pt}%
  \newcommand{\trheadh}{8pt}%
  \newcommand{\trfont}{\scriptsize}%
  \newcommand{\trlegfont}{\scriptsize}%
  \newcommand{\trlegline}{8pt}%
  \newcommand{\trpromptfont}{\scriptsize}%
  \newcommand{\trpromptw}{0.18\linewidth}%
  \newcommand{\trpromptfill}{black!2.5}%
  \definecolor{trbase}{HTML}{EB6834}%
  \definecolor{trours}{HTML}{2A78D6}%
  \definecolor{trrmg}{HTML}{4E9F5B}%
  \newcommand{\trv}[1]{\csname trajc@#1\endcsname}%
  \tikzset{trimg/.style={inner sep=\trpad, line width=\trframe},
           trdrop/.style={black!45, line width=0.4pt, dash pattern=on 1.5pt off 1.2pt},
           trdot/.style={only marks, mark=*, mark size=1.6pt, mark options={draw=white, line width=0.5pt}},
           trhead/.style={anchor=south, inner sep=0pt, font=\trfont, text=black!70}}%
  \begin{tikzpicture}
    \begin{axis}[name=ax, set layers, width=\trw, height=\trh, scale only axis, clip=false,
        xmin=2, xmax=\trxmax, ymin=\trymin, ymax=\trymax,
        xtick={2,4,8,20,50}, ytick={25,50,75,100}, minor xtick={}, tick pos=left,
        ymajorgrids, grid style={black!8, line width=0.3pt},
        tick label style={font=\trfont}, label style={font=\trfont},
        xlabel={Steps}, ylabel={FID-5k $\downarrow$}, xlabel style={at={(axis description cs:0.70,0)}, anchor=north, yshift=0pt}, ylabel style={yshift=-4pt}]
      \edef\trplot{\noexpand\addplot[trbase, line width=1.1pt] coordinates {\trv{base@curve}};
                   \noexpand\addplot[trrmg, line width=1.1pt] coordinates {\trv{rmg@curve}};
                   \noexpand\addplot[trours, line width=1.1pt] coordinates {\trv{rsd@curve}};
                   \noexpand\addplot[trdot, mark options={fill=trbase}] coordinates
                     {(4,\trv{base@4}) (8,\trv{base@8}) (20,\trv{base@20}) (50,\trv{base@50})};
                   \noexpand\addplot[trdot, mark options={fill=trrmg}] coordinates
                     {(4,\trv{rmg@4}) (8,\trv{rmg@8}) (20,\trv{rmg@20})};
                   \noexpand\addplot[trdot, mark options={fill=trours}] coordinates
                     {(4,\trv{rsd@4}) (8,\trv{rsd@8})};}\trplot
      \draw[trdrop, on layer=axis grid] (axis cs:4,\trymin) -- (axis cs:4,\trv{rmg@4});
      \draw[trdrop, on layer=axis grid] (axis cs:8,\trymin) -- (axis cs:8,\trv{rmg@8});
      \draw[trdrop, on layer=axis grid] (axis cs:20,\trymin) -- (axis cs:20,\trv{base@20});
      \draw[trdrop, on layer=axis grid] (axis cs:50,\trymin) -- (axis cs:50,\trv{base@50});
    \end{axis}
    \node[trimg, draw=trbase, anchor=north east] (b50) at ($(ax.north east)+(-\trinset,-\trinset-\trheadh)$)
      {\includegraphics[width=\trthumb]{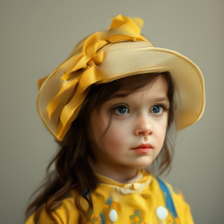}};
    \node[trimg, draw=trbase, anchor=east] (b20) at ($(b50.west)+(-\trcolgap,0)$)
      {\includegraphics[width=\trthumb]{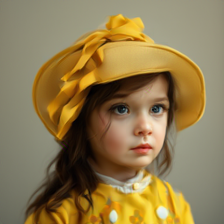}};
    \node[trimg, draw=trbase, anchor=east] (b8) at ($(b20.west)+(-\trcolgap,0)$)
      {\includegraphics[width=\trthumb]{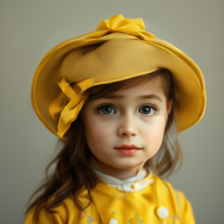}};
    \node[trimg, draw=trbase, anchor=east] (b4) at ($(b8.west)+(-\trcolgap,0)$)
      {\includegraphics[width=\trthumb]{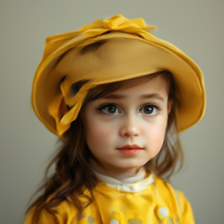}};
    \foreach \k/\f in {4/04, 8/08, 20/20} {
      \node[trimg, draw=trrmg, anchor=north] (m\k) at ($(b\k.south)+(0,-\trrowgap)$)
        {\includegraphics[width=\trthumb]{\trdir/rmg_nfe\f.png}};}
    \foreach \k/\f in {4/04, 8/08} {
      \node[trimg, draw=trours, anchor=north] (r\k) at ($(m\k.south)+(0,-\trrowgap)$)
        {\includegraphics[width=\trthumb]{\trdir/rsd_nfe\f.png}};}
    \foreach \k in {4,8,20,50}
      \node[trhead] at ($(b\k.north)+(0,1.5pt)$) {\k\ steps};
    \node[anchor=center, font=\trlegfont, align=left, inner sep=0pt]
      at ($(r8.north east)!0.5!(r8.south east -| b50.east)$)
      {\tikz[baseline=-0.6ex]\draw[trbase, line width=1.1pt] (0,0) -- (\trlegline,0);\ Baseline\\[1.5pt]
       \tikz[baseline=-0.6ex]\draw[trrmg, line width=1.1pt] (0,0) -- (\trlegline,0);\ RMG~\cite{curvo2026follow}\\[1.5pt]
       \tikz[baseline=-0.6ex]\draw[trours, line width=1.1pt] (0,0) -- (\trlegline,0);\ \textbf{\method} (ours)};
    \node[anchor=north west, inner xsep=4pt, inner ysep=2.5pt, font=\trpromptfont, fill=\trpromptfill, draw=black!10,
          line width=0.5pt, rounded corners=5pt, text width=\trpromptw, align=left]
      at ($(ax.north west)+(\trinset,-\trinset)$) {\hyphenpenalty=10000\relax\gridprompt{\trv{\trpromptkey}}};
  \end{tikzpicture}
  \caption{\textbf{A better start beats more steps.} FID-5k of Scale-RAE on COCO by number of
    sampling steps. In a few steps, \method beats the baseline and RMG~\cite{curvo2026follow} at any
    step count. Above: one WikiArt prompt, from the same seed.}
  \label{fig:traj_curve}
\end{figure}

This question leads to our method,
\emph{Retrieval-Seeded Diffusion} (\method),
which changes only the starting point
of a frozen semantic flow model (\cref{fig:teaser_traj}).
We keep a corpus of captioned images,
storing each image's caption embedding
in a text semantic space
and its latent in the model's image semantic space.
Then, given a prompt, we embed it in the same text space
and retrieve the latents of the images
whose captions lie nearest to it.
These latents are combined with a weighted average
to form an \emph{anchor},
an image latent that reflects the prompt.
With the anchor in hand, we shift the initial noise distribution a short step towards it,
to a Gaussian of the same form as the model's training distribution at an early noise level,
so the frozen model continues from it unchanged.
From this start, a few steps give a better FID than many steps from Gaussian noise (\cref{fig:traj_curve}).
Because the change lives entirely in the starting point,
\method trains nothing: the model's weights, schedule and conditioning are unchanged, and the only
addition is a store of corpus latents, which can be edited without retraining.
Each generation is also linked to the corpus images that seeded it,
which supports attributed generation~\cite{wang2023evaluating,georgiev2023journey}.

Our main contributions are: (1) we propose \method, a training-free way to give a frozen semantic
flow model a prompt-dependent start, by moving its initial noise a short step towards a
similarity-weighted mean of the latents of corpus images retrieved for the prompt; (2) on three
semantic flow models~\cite{tong2026scalerae,shi2025svgt2i,feng2026sefi} and three datasets, \method
lowers FID to the target dataset in eight of nine settings in $6.25\times$ fewer sampling steps, with
comparable prompt alignment, and matches or beats the FID of RMG~\cite{curvo2026follow} in every
setting with $2.5\times$ fewer steps; (3) we introduce the \emph{impurity} of a retrieval, the share
of the corpus that lies closer to the anchor than the retrieved neighbours do, and find it lower in
all three semantic spaces than in a VAE latent space, where a mean of neighbours describes the prompt
less well.

\section{Related work}
\label{sec:related}
\paragraph{Semantic latent spaces.}
Latent diffusion trains the generator in the latent space of an autoencoder fitted for pixel
reconstruction~\cite{rombach2021highresolution}. Several works bring that space closer to a
pretrained representation encoder, by aligning the denoiser's features~\cite{yu2025repa} or the
latent~\cite{yao2025vavae} with it, and report faster training. Representation autoencoders take a
frozen encoder's features as the latent space and train only a decoder~\cite{zheng2025rae}, and
SVG~\cite{shi2025svg} builds its latent from frozen DINOv3 features with a light residual encoder for
fine detail. Scale-RAE~\cite{tong2026scalerae} and SVG-T2I~\cite{shi2025svgt2i} scale these two designs
to text-to-image generation, over SigLIP-2 and DINOv3 features respectively, and
SeFi-Image~\cite{feng2026sefi} generates such a semantic latent before a texture one. These are the backbones we sample.
In a VAE's latent space images group by appearance, not content,
so we expect a smaller gain there, as we discuss later on.

\paragraph{Retrieval-based image generation.}
Retrieval augmentation, popularised in language models~\cite{lewis2020rag}, has been attached to
image generators at several points: conditioning a trained model on retrieved
neighbours~\cite{blattmann2022retrieval,chen2023reimagen,sheynin2023knn,shrestha2024fairrag,xin2026pixarg}, handing a frozen generator reference
images at inference~\cite{shalevarkushin2025imagerag}, or training the retriever against the
generator's failures~\cite{lyu2025realrag}. All of them change what the model sees, through its
conditioning or its context, while \method changes only where the sampler starts.
ImageRAGTurbo~\cite{qiu2026imageragturbo} also searches a text embedding space, but finetunes a
one-step backbone with a retrieval adapter, while \method trains nothing.
Closest to ours is Reference-Mean Guidance~\cite{curvo2026follow}, proposed in a VAE latent space,
which also trains nothing, builds a weighted mean of reference latents and changes the velocity by it
over most of the trajectory. We found no counterpart designed for semantic latent spaces like \method. \method is
$2.5\times$ faster than RMG (\cref{fig:traj_curve}), with a better FID on all three backbones and a
better KID on two at a similar CLIP score, and is more
diverse than the released sampler wherever RMG lowers diversity in some cases. PixARG~\cite{xin2026pixarg} starts from a noised
retrieved image but also trains its denoiser, while \method only sets the start and trains nothing.

\paragraph{Initial noise.}
Several methods change the initial noise per prompt and keep the model~\cite{samuel2023nao,samuel2024seedselect,li2026oracle,wang2026anchorsteer}. InitNO~\cite{guo2024initno}
optimises it at test time against the model's attention maps, \citet{ahn2024noise} train a refiner
whose noise can stand in for guidance,
and Golden Noise~\cite{zhou2024golden} and noise hypernetworks~\cite{eyring2025noise} train networks
that predict an improved noise from the prompt. A noise can also be retrieved from a
library of pre-generated ones~\cite{wang2025noisequery}, or shifted by the statistics of a target image
set to set its style~\cite{everaert2024signal}. SDEdit~\cite{meng2022sdedit} noises a user-supplied
image to a level that trades faithfulness against realism, and noising a good initial estimate also
cuts the steps needed~\cite{chung2022come,lyu2022esddpm}. Our initial state has the same form, but its
mean is an estimate of where the flow's first step goes, rather
than an input to preserve. To our knowledge, \method is the only method that moves
the initial noise per prompt towards real images without training, test-time optimisation or extra
network evaluations.

\paragraph{Source distributions for flow matching.}
A second line changes the source distribution itself~\cite{liu2025crossflow} or its coupling to the
data~\cite{pooladian2023multisample,albergo2024couplings}. \citet{lee2025source} ask which source a flow
model should be trained with, and \citet{malnick2026otfm} pair each image with a low-frequency
version of itself so that the coupling is near optimal by construction, sampling that source at
inference with a small auxiliary model. Conditional priors make the source depend on the condition~\cite{scholz2025warm,chen2025carflow}:
\citet{issachar2025designing} centre it on an average of the data that share the condition and report
fewer sampling steps, and \citet{kim2026bettersourcebetterflow} learn it under the flow-matching
objective. These are the closest to our anchor in form, a condition-specific centre for the source,
but all are fixed during training, while ours displaces the source of a model trained with the
Gaussian one, per prompt and at sampling time.
\citet{kim2026bettersourcebetterflow} also report larger gains in an RAE latent space than in an
SD-VAE one.

\paragraph{Few-step sampling and diversity.}
Training-free solvers and schedules~\cite{song2021ddim,lu2022dpmsolver,sabour2024ays} change how
the trajectory is integrated, and caching methods reuse a stored state of an earlier generation with
a similar prompt or early trajectory~\cite{agarwal2024nirvana,lu2024recon,zhang2023redi}. \method keeps the released
scheduler.
For a finite training set, the optimal velocity is a weighted mean of
the training latents, and following it at every step should reproduce training
images~\cite{scarvelis2023closed,kamb2025analytic}. \method uses a weighted mean of this kind for
the first step only. Other methods reduce the number of steps through training, by
distillation~\cite{salimans2022progressive,song2023consistency,luo2023lcm,sauer2023add,yin2024dmd2}
or by learning average velocities~\cite{geng2025meanflow}. Per-prompt diversity can be raised by
annealing the condition~\cite{sadat2024cads} or by optimising the noise~\cite{shi2026mono}. Since an
anchor might lower it, we measure it with the protocol of the latter.

\section{Preliminaries}
\label{sec:preliminaries}
\begin{figure*}[!t]
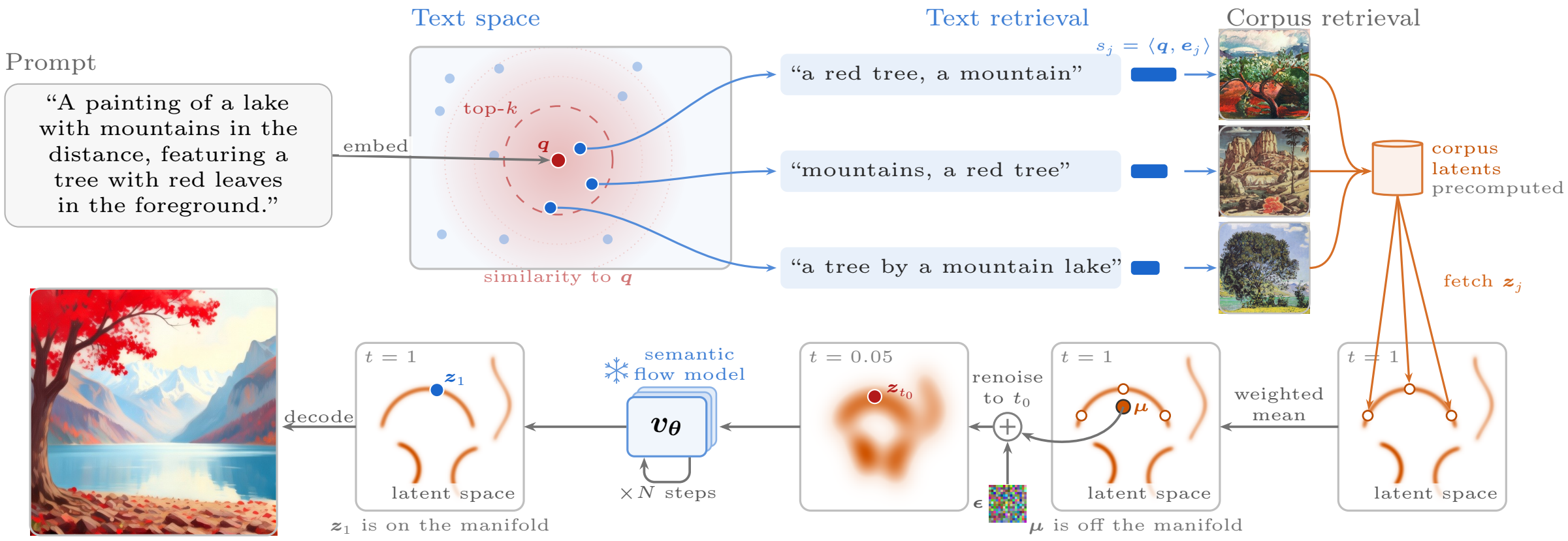

  \centering
  \input{figures/method/coords}%
  \newcommand{\mfigimg}{figures/method/img}
  \newcommand{\figfont}{\usefont{T1}{SourceSansPro-LF}{m}{n}}
  \newcommand{\figbold}{\usefont{T1}{SourceSansPro-LF}{sb}{n}}
  \newcommand{\promptfont}{\usefont{T1}{comfortaa}{m}{n}}
  \newcommand{\mfigthumb}[3]{%
    \IfFileExists{#3}{%
      \begin{scope}
        \clip ($#1-(#2/2,#2/2)$) rectangle ($#1+(#2/2,#2/2)$);
        \node[inner sep=0pt] at #1 {\includegraphics[height=#2]{#3}};
      \end{scope}}{%
      \fill[black!10] ($#1-(#2/2,#2/2)$) rectangle ($#1+(#2/2,#2/2)$);}}
  \newcommand{\mfigfrost}[2]{%
    \begin{scope}[shift={#1}, draw=otherblue!70, line width=0.35pt, line cap=round]
      \foreach \a in {30,90,150,210,270,330}{
        \draw (0,0) -- (\a:#2);
        \draw (\a:0.62*#2) -- ++(\a+40:0.30*#2);
        \draw (\a:0.62*#2) -- ++(\a-40:0.30*#2);}
    \end{scope}}
  \ifdefined\figcachebegin\figcachebegin{method}\fi
  \begin{tikzpicture}[
      font=\figfont\scriptsize,
      arr/.style={-{Stealth[length=3.5pt]}, line width=0.7pt, black!55},
      txt arr/.style={-{Stealth[length=3pt]}, line width=0.6pt, otherblue!75, rounded corners=3pt},
      lat arr/.style={-{Stealth[length=3pt]}, line width=0.6pt, burntorange!85, rounded corners=3pt},
      ttl/.style={font=\figbold\scriptsize, inner sep=0pt},
      note/.style={font=\figfont\tiny, text=black!65, inner sep=1pt},
      dot/.style={circle, draw=white, line width=0.4pt, inner sep=1.4pt},
      zdot/.style={circle, fill=white, draw=burntorange!90!black, line width=0.5pt, inner sep=1pt},
      mudot/.style={circle, fill=burntorange, draw=black!80, line width=0.5pt, inner sep=1.4pt},
      panel/.style={draw=black!25, line width=0.6pt, rounded corners=3pt},
      plabel/.style={note, anchor=north west, fill=white, fill opacity=0.8, text opacity=1, rounded corners=1pt},
      halo/.style={fill=white, fill opacity=0.8, text opacity=1, inner sep=0.6pt, rounded corners=1pt},
    ]
    \def\yA{3.35}%
    \def\yB{0.65}%
    \def\S{1.7}

    \begin{scope}[yshift=0.3cm]%
    \def\yT{4.42}
    \node[draw=black!30, fill=black!3, rounded corners=4pt, text width=3.1cm, align=center,
          font=\promptfont\fontsize{6.3}{7.4}\selectfont, inner sep=3pt] (prompt) at (1.85,\yA-0.25)%
          {``A painting of a lake with mountains in the distance, featuring a tree with red leaves in the foreground.''};
    \node[ttl, text=black!60, anchor=south west] at ($(prompt.north west)+(0,0.08)$) {Prompt};

    \begin{scope}%
      \clip[rounded corners=3pt] (4.3,1.95) rectangle (7.55,4.2);
      \fill[otherblue!4] (4.3,1.95) rectangle (7.55,4.2);
      \shade[inner color=darkred!30, outer color=otherblue!4] (5.80,3.05) circle (1.3);
      \foreach \r in {0.85,1.15} \draw[darkred!22, line width=0.35pt, densely dotted] (5.80,3.05) circle (\r);
    \end{scope}
    \draw[panel] (4.3,1.95) rectangle (7.55,4.2);
    \node[ttl, text=otherblue!85, anchor=base west] at (4.3,\yT) {Text space};
    \foreach \x/\y in {4.62/2.30, 4.60/3.55, 4.70/3.88, 5.25/2.25, 5.20/3.98,
                       6.30/2.25, 6.60/4.00, 5.15/3.10, 6.45/3.70}
      \node[circle, fill=otherblue!55, inner sep=1pt, opacity=0.6] at (\x,\y) {};
    \draw[darkred!65, dashed, line width=0.5pt] (5.80,3.05) circle (0.55);
    \node[circle, fill=otherblue, draw=white, line width=0.4pt, inner sep=1.3pt] (h1) at (6.02,3.17) {};
    \node[circle, fill=otherblue, draw=white, line width=0.4pt, inner sep=1.3pt] (h2) at (6.14,2.81) {};
    \node[circle, fill=otherblue, draw=white, line width=0.4pt, inner sep=1.3pt] (h3) at (5.72,2.57) {};
    \node[circle, fill=darkred, draw=white, line width=0.4pt, inner sep=1.5pt] (q) at (5.80,3.05) {};
    \node[note, text=darkred, anchor=south east] at (5.76,3.09) {$\vq$};
    \node[note, text=darkred!85, anchor=south east] at (5.44,3.45) {top-$k$};
    \node[note, text=darkred!60, anchor=north, font=\figfont\tiny\itshape] at (5.80,1.98) {similarity to $\vq$};
    \draw[arr] (prompt.east) -- node[note, above, pos=0.20, fill=white, fill opacity=0.85, text opacity=1] {embed} (q);

    \node[ttl, text=otherblue!85, anchor=base] at (10.35,\yT) {Text retrieval};
    \node[ttl, text=black!60, anchor=base] at (13.55,\yT) {Corpus retrieval};
    \node[text=otherblue!85, anchor=south, inner sep=1pt,
          font=\figfont\fontsize{4.8}{5.2}\selectfont] at (11.83,4.07)
          {$s_j=\langle\vq,\ve_j\rangle$};
    \foreach \c/\s/\sz/\hz/\y [count=\i] in {{a red tree, a mountain}/0.92/0.92/0.46/3.92,
                                            {mountains, a red tree}/0.74/0.92/0.46/2.94,
                                            a tree by a mountain lake/0.58/0.92/0.46/1.96}{
      \fill[otherblue, fill opacity=0.1, rounded corners=2pt] (8.05,\y-0.21) rectangle (11.5,\y+0.21);
      \node[note, anchor=west, text=black!85, font=\promptfont\fontsize{5.8}{6.4}\selectfont] at (8.12,\y) {``\c''};
      \fill[otherblue, rounded corners=1pt] (11.6,\y-0.07) rectangle ++(\s*0.50,0.14);
      \mfigthumb{(12.95,\y)}{\sz cm}{\mfigimg/nb\i.jpg}
      \draw[panel, line width=0.5pt] (12.95-\hz,\y-\hz) rectangle (12.95+\hz,\y+\hz);
      \coordinate (th\i) at (12.95+\hz,\y);%
      \draw[txt arr] (12.14,\y) -- (12.93-\hz,\y);}
    \draw[txt arr] (h1) to[out=0, in=180] (8.01,3.92);
    \draw[txt arr] (h2) to[out=0, in=180] (8.01,2.94);
    \draw[txt arr] (h3) to[out=0, in=180] (8.01,1.96);
    \end{scope}

    \coordinate (PA) at (14.55,\yB);
    \coordinate (PW) at (11.65,\yB);
    \coordinate (PU) at (9.10,\yB);
    \coordinate (P2) at (4.60,\yB);
    \foreach \c in {PA,PW,P2}
      \node[inner sep=0pt] at (\c) {\includegraphics[width=\S cm]{\mfigimg/marg_data}};
    \node[inner sep=0pt] at (PU) {\includegraphics[width=\S cm]{\mfigimg/marg_u0}};
    \draw[panel] ($(PU)-(\S/2,\S/2)$) rectangle ($(PU)+(\S/2,\S/2)$);
    \foreach \c in {PA,PW,P2}
      \draw[panel] ($(\c)-(\S/2,\S/2)$) rectangle ($(\c)+(\S/2,\S/2)$)
        node[plabel, anchor=south east] at ($(\c)+(\S/2-0.04,-\S/2+0.04)$) {latent space};
    \foreach \c/\t in {PA/1, PW/1, PU/0.05, P2/1}
      \node[plabel, text=black!50] at ($(\c)+(-\S/2+0.04,\S/2-0.04)$) {$t = \t$};
    \begin{scope}[shift={(PA)}, x=\S cm/4, y=\S cm/4]
      \foreach \k in {ZjA,ZjB,ZjC} \node[zdot] (\k) at (\csname mfig\k x\endcsname,\csname mfig\k y\endcsname) {};
    \end{scope}
    \begin{scope}[shift={(PW)}, x=\S cm/4, y=\S cm/4]
      \foreach \k in {ZjA,ZjB,ZjC} \node[zdot] (w\k) at (\csname mfig\k x\endcsname,\csname mfig\k y\endcsname) {};
      \coordinate (mu) at (\mfigMux,\mfigMuy);
    \end{scope}
    \begin{scope}[shift={(PU)}, x=\S cm/4, y=\S cm/4] \coordinate (z0) at (\mfigZzerox,\mfigZzeroy); \end{scope}
    \begin{scope}[shift={(P2)}, x=\S cm/4, y=\S cm/4]
      \coordinate (z1) at (\mfigZonex,\mfigZoney);
    \end{scope}

    \node[cylinder, shape border rotate=90, aspect=0.25, draw=burntorange!80, fill=burntorange!8,
          line width=0.6pt, minimum width=0.5cm, minimum height=0.55cm] (store) at (14.30,3.24) {};
    \node[note, anchor=west, align=left, text=burntorange!95!black] at ($(store.east)+(0.04,0)$) {corpus\\latents\\[-0.5pt]\textit{\color{black!55}precomputed}};
    \foreach \i in {1,2,3} \draw[lat arr] (th\i) to[out=0, in=180] (store.west);
    \foreach \k in {ZjA,ZjB,ZjC} \draw[lat arr] (store.south) -- (\k);
    \node[note, text=burntorange!90, anchor=west, font=\figfont\tiny\itshape] at (14.72,2.10) {fetch $\vz_j$};

    \draw[arr] (13.70,\yB) -- node[note, above, align=center, inner sep=1.5pt] {weighted\\mean} (12.50,\yB);
    \node[mudot] at (mu) {};
    \node[note, halo, text=burntorange!95!black, anchor=west] at ($(mu)+(0.07,-0.02)$) {$\vmu$};
    \node[note, text=black!55, anchor=north, font=\figfont\tiny\itshape] at ($(PW)+(0,-\S/2-0.04)$) {$\vmu$ is off the manifold};

    \node[circle, draw=black!50, line width=0.6pt, inner sep=0pt, minimum size=0.28cm] (plus) at (10.35,\yB) {};
    \draw[black!50, line width=0.6pt] ($(plus)+(-0.08,0)$) -- ($(plus)+(0.08,0)$) ($(plus)+(0,-0.08)$) -- ($(plus)+(0,0.08)$);
    \draw[arr] (mu) .. controls (11.40,0.45) and (10.80,0.45) .. (plus);%
    \node[inner sep=0pt] (eps) at (10.35,\yB-0.78) {\includegraphics[width=0.38cm]{\mfigimg/noise}};
    \node[note, anchor=east] at (eps.west) {$\veps$};
    \draw[arr] (eps) -- (plus);
    \node[note, anchor=south, align=center] at ($(plus)+(0,0.16)$) {renoise\\to $t_0$};
    \draw[arr] (plus) -- (9.95,\yB);

    \node[dot, fill=darkred] at (z0) {};
    \node[note, text=darkred, anchor=west] at ($(z0)+(0.07,0.05)$) {$\vz_{t_0}$};

    \coordinate (M) at (6.9,\yB);
    \foreach \o/\t in {0.1/12, 0.05/18} \draw[fill=otherblue!\t, draw=otherblue!50, rounded corners=2pt, line width=0.5pt]
       ($(M)+(-0.4+\o,-0.3+\o)$) rectangle ($(M)+(0.4+\o,0.3+\o)$);
    \node[draw=otherblue!70, fill=otherblue!6, rounded corners=2pt, line width=0.6pt,
          minimum width=0.8cm, minimum height=0.6cm, font=\small] (blk) at (M) {$\vv_{\vtheta}$};
    \draw[arr, rounded corners=3pt] ($(blk.south east)+(-0.18,0)$) -- ++(0,-0.25) -| ($(blk.south west)+(0.18,0)$);
    \node[note, anchor=north] at ($(blk.south)+(0,-0.25)$) {$\times N$ steps};
    \node[note, anchor=south, align=center, font=\figbold\tiny, text=otherblue!85] at ($(blk.north)+(0.23,0.12)$) {semantic\\flow model};
    \mfigfrost{($(blk.north west)+(-0.10,0.26)$)}{0.13cm}
    \draw[arr] (8.25,\yB) -- ($(blk.east)+(0.12,0)$);
    \draw[arr] (blk.west) -- (5.45,\yB);

    \node[dot, fill=otherblue] at (z1) {};
    \node[note, halo, text=otherblue, anchor=south west] at ($(z1)+(0.04,0.02)$) {$\vz_1$};
    \node[note, text=black!55, anchor=north, font=\figfont\tiny\itshape] at ($(P2)+(0,-\S/2-0.04)$) {$\vz_1$ is on the manifold};

    \coordinate (O) at (1.70,\yB+0.15);%
    \mfigthumb{(O)}{2.5cm}{\mfigimg/out.jpg}%
    \draw[panel] ($(O)-(1.25,1.25)$) rectangle ($(O)+(1.25,1.25)$);
    \draw[arr] (3.75,\yB) -- node[note, above] {decode} (3.00,\yB);
  \end{tikzpicture}
  \ifdefined\figcacheend\figcacheend\fi
  \caption{\textbf{Method overview.} We retrieve top-k images by caption similarity, then compute the anchor from the retrieved semantic image latents that are matched with those captions. We then use the anchor to shift the initial seed of the generation process, and the frozen model runs $N$ steps. Nothing is trained.}
  \label{fig:method}
\end{figure*}

Our setting is a pretrained, frozen text-to-image flow model that generates in a semantic latent
space $\R^D$. Time along the generative path is a scalar $t \in [0, 1]$, with $t = 0$ pure noise and
$t = 1$ a real image latent. The condition $\vc$ is the prompt's embedding under the backbone's own
text encoder, fixed for the whole trajectory, and the query $\vq$ is its embedding under a separate
retrieval encoder, renormalised to the unit sphere. We write $\veps \sim \N(\vzero, \mI)$ for
Gaussian noise in latent space. A subscript in $t$ indexes time and a subscript in $j$ the corpus.

When sampling a pretrained flow model, the network $\vv_\vtheta(\vz_t, t, \vc)$
is trained to predict the velocity along the linear path
between noise and data~\cite{lipman2023flow,liu2023rectified},
\begin{equation}
  \vz_t = t \vz_1 + (1 - t)\, \veps, \qquad \veps \sim \N(\vzero, \mI),
  \label{eq:interp}
\end{equation}
and sampling integrates the ordinary differential equation
\begin{equation}
  \frac{d\vz_t}{dt} = \vv_\vtheta(\vz_t, t, \vc)
  \label{eq:ode}
\end{equation}
from $t = 0$ to $t = 1$. In practice this takes $N$ Euler steps,
\begin{equation}
  \vz_{t_{i+1}} = \vz_{t_i} + (t_{i+1} - t_i)\, \vv_\vtheta(\vz_{t_i}, t_i, \vc),
  \label{eq:euler}
\end{equation}
with $t_0 < t_1 < \dots < t_N = 1$, from $\vz_{t_0} \sim \N(\vzero, \mI)$ at $t_0 = 0$. The time
points are $N + 1$ points of the model's released grid on $[t_0, 1]$, both endpoints included, so a
shorter budget visits fewer of the model's own points rather than a schedule of ours.
Classifier-free guidance~\cite{ho2022classifier} evaluates the velocity with the condition $\vc$ and
with a null condition $\varnothing$ and extrapolates with a scale $w$,
\begin{equation}
  \tilde\vv = \vv_\vtheta(\vz, t, \varnothing) + w \big( \vv_\vtheta(\vz, t, \vc) - \vv_\vtheta(\vz, t, \varnothing) \big).
  \label{eq:cfg}
\end{equation}
At $w = 1$ this is plain conditional sampling. Larger $w$ increases prompt adherence at the cost of
per-prompt diversity~\cite{sadat2024cads}. A step is one
forward pass at $w \in \{0, 1\}$ and two otherwise, which can run in parallel.

\paragraph{Retrieval.}
A corpus $\gD = \{(\ve_j, \vz_j)\}_{j=1}^{M}$ is indexed once, offline. The key $\ve_j$ is the mean of
the image's caption embeddings under the retrieval encoder, renormalised to the unit sphere, so query
and key sit on the same side of the encoder's modality gap and the search compares text to text. The
latent $\vz_j \in \R^D$ is the image under the backbone being sampled, which is what generation
starts from. At sampling time we write $\gI_k(\vq)$ for the indices of the $k$ keys of largest cosine
similarity $s_j = \langle \vq, \ve_j \rangle$, and $\mZ_{\vq}(k) = \{\vz_j\}_{j \in \gI_k(\vq)} \in
\R^{k \times D}$ for their latents. The search is exact, a single matrix product, though an
approximate nearest-neighbour index~\cite{johnson2019faiss} would help for a much
larger corpus.

\section{Retrieval-Seeded Diffusion (\method)}
\label{sec:method}
This section presents \method (\cref{fig:method}), a training-free way to give a frozen semantic flow
model a prompt-dependent start. We average the latents of the corpus images retrieved for the prompt
into an anchor, shift the initial noise towards it, and let the frozen model sample from there.
Specifically, given a text prompt, we retrieve $k$ corpus latents $\mZ_{\vq}(k)$ and their similarities $s_j$ to
the prompt as described in \cref{sec:preliminaries}, and combine them into a single latent, the
anchor
\begin{equation}
  \vmu = \sum_{j \in \gI_k(\vq)} \alpha_j \vz_j,
  \qquad
  \alpha_j = \frac{\exp(s_j / \tau)}{\sum_{l \in \gI_k(\vq)} \exp(s_l / \tau)} ,
  \label{eq:anchor}
\end{equation}
a similarity-weighted mean whose temperature $\tau$ sets how the weight spreads over the neighbours,
from the nearest neighbour alone ($\tau \to 0$) to a plain mean over the $k$ neighbours ($\tau \to \infty$).
We set $\tau = 0.02$ throughout, and
\cref{sec:exp:diversity} and the ablations in \cref{app:ablations} report what the two limits of $\tau$ cost.

\begin{figure}[t]
  \centering
  \input{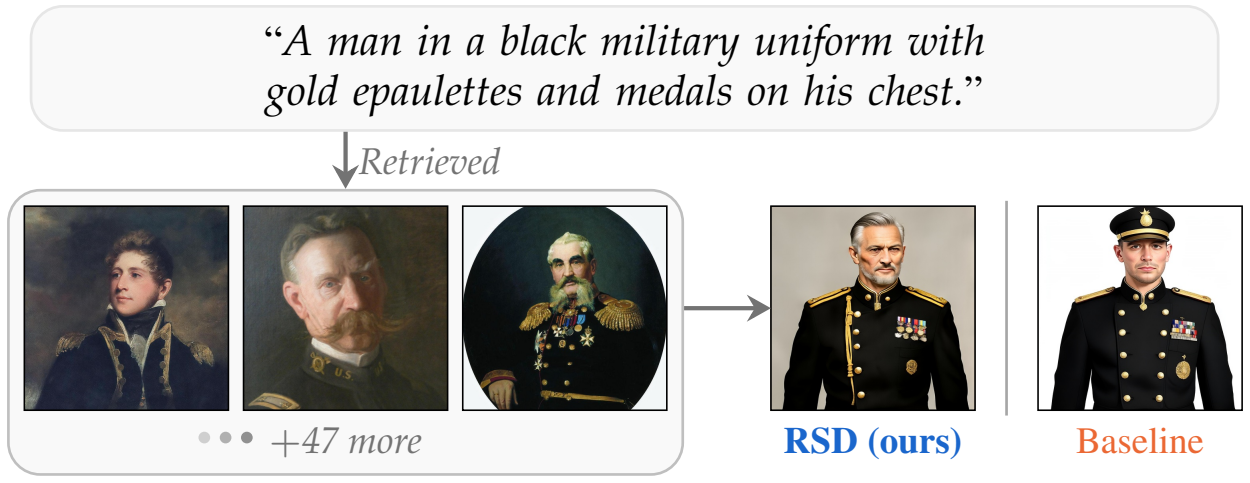}%
  \newcommand{\nniside}{0.162\linewidth}%
  \newcommand{\nnigap}{0.012\linewidth}%
  \newcommand{\nniarrow}{0.07\linewidth}%
  \newcommand{\nnisep}{0.05\linewidth}%
  \newcommand{\nniretpad}{3pt}%
  \newcommand{\nnilabsep}{2pt}%
  \newcommand{\nnipromptlift}{14pt}%
  \newcommand{\nniframe}{0.3pt}%
  \newcommand{\nnifont}{\scriptsize}%
  \newcommand{\nnipctfont}{\usefont{T1}{ppl}{m}{it}}%
  \newcommand{\nnipromptfont}{\footnotesize}%
  \newcommand{\nnipromptw}{0.9\linewidth}%
  \newcommand{\nnifill}{black!2.5}%
  \newcommand{\nniretdraw}{black!25}%
  \newcommand{\nnidotsep}{4pt}%
  \definecolor{nnibase}{HTML}{EB6834}%
  \newcommand{\nniimg}[1]{\includegraphics[width=\nniside,height=\nniside]{figures/nn_influence/img/#1}}%
  \ifdefined\figcachebegin\figcachebegin{nn_influence}\fi
  \begin{tikzpicture}[panel/.style={inner sep=0pt, outer sep=0pt},
                      pct/.style={inner sep=0pt, font=\nnifont\nnipctfont},
                      flow/.style={black!55, line width=0.8pt, -{Stealth[length=4.5pt, width=5pt]}}]
    \node[panel, anchor=south west] (n1) at (0,0) {\nniimg{nb1.jpg}};
    \node[panel, anchor=west] (n2) at ($(n1.east)+(\nnigap,0)$) {\nniimg{nb2.jpg}};
    \node[panel, anchor=west] (n3) at ($(n2.east)+(\nnigap,0)$) {\nniimg{nb3.jpg}};
    \node[pct, anchor=north, text=black!55] (more) at ($(n2.south)+(0,-\nnilabsep-1pt)$) {$+$\nnirest\ more};
    \foreach \d in {1,2,3}
      \fill[black!\the\numexpr 55-12*\d\relax] ($(more.west)+(-\d*\nnidotsep+0.5*\nnidotsep-2pt,0.2ex)$) circle (1.1pt);
    \begin{scope}[on background layer]
      \node[fit=(n1)(n2)(n3)(more), inner sep=\nniretpad, fill=\nnifill, draw=\nniretdraw, line width=0.5pt,
            rounded corners=6pt] (ret) {};
    \end{scope}
    \node[panel, anchor=west] (ours) at ($(n3.east)+(\nniretpad+\nniarrow,0)$) {\nniimg{ours.jpg}};
    \node[panel, anchor=west] (base) at ($(ours.east)+(\nnisep,0)$) {\nniimg{base.jpg}};
    \foreach \n in {n1,n2,n3,ours,base}
      \draw[black, line width=\nniframe] (\n.south west) rectangle (\n.north east);
    \draw[flow] (n3.east -| ret.east) -- (ours.west);
    \draw[black!35, line width=0.5pt] ($(ours.south east)!0.5!(base.south west)+(0,-1pt)$)
                                   -- ($(ours.north east)!0.5!(base.north west)+(0,1pt)$);
    \node[anchor=north, inner sep=0pt, font=\nnifont] at ($(ours.south)+(0,-\nnilabsep-1pt)$)
      {\textcolor{otherblue}{\textbf{\method{} (ours)}}};
    \node[anchor=north, inner sep=0pt, font=\nnifont] at ($(base.south)+(0,-\nnilabsep-1pt)$) {\textcolor{nnibase}{Baseline}};
    \path let \p1 = (ret.west), \p2 = (base.east), \p3 = (n1.north) in%
      node[inner ysep=2.5pt, inner xsep=5pt, anchor=south, font=\nnipromptfont, fill=\nnifill, draw=black!10,
           line width=0.5pt, rounded corners=6pt, text width=\nnipromptw, align=center]
        (prompt) at ({0.5*(\x1+\x2)}, {\y3+\nnipromptlift}) {\gridprompt{\nniprompt}};
    \draw[flow] (prompt.south -| ret.north) -- node[right, pct, text=black!55, inner sep=2pt] {Retrieved} (ret.north);
  \end{tikzpicture}%
  \ifdefined\figcacheend\figcacheend\fi
  \caption{\textbf{The domain shapes the output.} \method's output takes on attributes shared by the
  paintings retrieved for the prompt, here a bare head and the background, which the released
  sampler does not follow.}
  \label{fig:nn_influence}
\end{figure}

Sampling starts from pure noise, $\vz_0 = \veps$, drawn independently of the image.
A flow model is trained to predict
the direction from the noise to the image at every stage, $\vz_1 - \veps$. Since it cannot know which image will come
out, the best it can learn is the average direction over the images that fit the
prompt, $\vv^\star(\vz_t, t, \vc) = \E[\vz_1 - \veps \mid \vz_t, \vc]$, the minimiser of the flow
matching loss~\cite{lipman2023flow,liu2023rectified}. At $t = 0$ the state is pure noise and tells
nothing about the image, so this average reduces to
\begin{equation}
  \vv^\star(\veps, 0, \vc) = \E[\vz_1 \mid \vc] - \veps ,
  \label{eq:vstar}
\end{equation}
where $\E[\vz_1 \mid \vc]$ is the mean latent of the images, given the prompt. The frozen model is
trained to predict exactly this direction at its first step, and a first Euler step of size $t_0$ along
it lands at $\veps + t_0 \vv^\star = t_0\, \E[\vz_1 \mid \vc] + (1 - t_0)\, \veps$.
The model has to recover this mean from its weights.
Instead, we estimate it from real images: the
corpus images whose captions match the prompt are examples of what the prompt describes, so the mean
of their latents estimates $\E[\vz_1 \mid \vc]$ directly. This is the anchor $\vmu$ of
\cref{eq:anchor}, a weighted average of the retrieved latents in which an image counts more the closer
its caption is to the prompt. Averaging examples by how similar their conditions are is the classical
kernel estimate of a conditional mean~\cite{nadaraya1964estimating,watson1964smooth}, with $\tau$ as
the squared bandwidth. \method takes the first step with $\vmu$ in place of $\E[\vz_1 \mid \vc]$ at no network
evaluation, and the remaining $N$ steps with the frozen model from $t_0$ to $1$. Sampling
therefore starts from
\begin{equation}
  \vz_{t_0} = t_0 \vmu + (1 - t_0)\, \veps \;\sim\; \N\big( t_0 \vmu, \; (1 - t_0)^2 \mI \big),
  \label{eq:state}
\end{equation}
ordinary Gaussian noise, centred at $t_0 \vmu$ instead of $\vzero$ and slightly narrower.
At our default, $t_0 = 0.05$, the state is $95\%$ noise, and from $t_0$ on the
model's own velocity is free to steer it away from $\vmu$. This short step is nonetheless enough for the retrieved images to
shape the output (\cref{fig:nn_influence}). It also lets \method at $8$ steps reach a lower FID than
Gaussian sampling at $50$ in almost every case we test. The anchor is deterministic given
the prompt, as the velocity of \cref{eq:vstar} is, so every sample from a prompt differs only through
$\veps$, exactly as in ordinary diffusion.
The $N$ steps then follow the model's released time grid on $[t_0, 1]$, so
the schedule is the model's own, started at $t_0$ instead of $0$.

\paragraph{Cost.}
\label{sec:method:cost}
Per prompt, \method adds one text embedding, one nearest-neighbour search and a read of $k$ latents
from a precomputed store. None of this calls the flow model, so it adds no step, and the search and the
read of $k = 50$ latents each cost much less than one forward pass (see \cref{tab:latency}).
The cost specific to \method is the store. The corpus latents must be kept in the backbone's own latent
space, where the anchor is computed, which takes up to tens of GiB in our experiments and
the keys take $\sim$300\,MiB (see \cref{tab:store_size}).
The store is built once and read for every prompt.
A network that predicts the anchor from the prompt can replace it.
We train one and find its results
close to those of retrieval (see \cref{app:learned_anchor}),
but it must be retrained for every new corpus.

\section{Experiments}
\label{sec:experiments}

\subsection{Setup}
\label{sec:exp:setup}

\begin{table*}[t]
\centering
\small
\setlength{\tabcolsep}{3.1pt}
\renewcommand{\arraystretch}{1.2}
\setlength{\aboverulesep}{0pt}
\setlength{\belowrulesep}{0pt}
\caption{\textbf{\method against each backbone's released sampler and RMG.} FID-5k, KID and CLIP on the $5{,}000$ held-out prompts of each dataset. Speedup is in sampling steps relative to the released sampler's $50$. On Scale-RAE the baseline is with $w=1$ and therefore RSD's speedup relative to it in NFE terms is $\times$3.125, we report steps speedup because model calls at each step with $w>1$ can be parallelized.
}
\label{tab:main}
\begin{tabular}{lc!{\color{black!45}\vrule width 0.25pt}ccc!{\color{black!45}\vrule width 0.25pt}ccc!{\color{black!45}\vrule width 0.25pt}ccc}
\toprule
 & \multicolumn{1}{c!{\color{black!45}\vrule width 0.25pt}}{} & \multicolumn{3}{c!{\color{black!45}\vrule width 0.25pt}}{COCO} & \multicolumn{3}{c!{\color{black!45}\vrule width 0.25pt}}{Anime} & \multicolumn{3}{c}{WikiArt} \\
\cmidrule(lr){3-5} \cmidrule(lr){6-8} \cmidrule(lr){9-11}
Method & Speedup$\uparrow$ & FID$\downarrow$ & KID{\scriptsize$\times 10^3$}$\downarrow$ & CLIP$\uparrow$ & FID$\downarrow$ & KID{\scriptsize$\times 10^3$}$\downarrow$ & CLIP$\uparrow$ & FID$\downarrow$ & KID{\scriptsize$\times 10^3$}$\downarrow$ & CLIP$\uparrow$ \\
\midrule
\multicolumn{2}{l!{\color{black!45}\vrule width 0.25pt}}{\textit{Scale-RAE}} & & & & & & & & & \\
Baseline & $\times$1 & 29.51 & 7.77 & 0.418 & 62.56 & 50.75 & 0.451 & 74.34 & 40.98 & 0.440 \\
+ RMG~\cite{curvo2026follow} & $\times$2.5 & 29.36 & 6.93 & \textbf{0.436} & 68.06 & 50.55 & 0.363 & 52.50 & 24.76 & \textbf{0.469} \\
\cellcolor{gray!10}+ \method{} (ours) & \cellcolor{gray!10}\textbf{\boldmath$\times$6.25} & \cellcolor{gray!10}\textbf{21.06} & \cellcolor{gray!10}\textbf{3.20} & \cellcolor{gray!10}0.420 & \cellcolor{gray!10}\textbf{49.84} & \cellcolor{gray!10}\textbf{42.36} & \cellcolor{gray!10}\textbf{0.467} & \cellcolor{gray!10}\textbf{32.07} & \cellcolor{gray!10}\textbf{10.19} & \cellcolor{gray!10}0.458 \\
\midrule
\multicolumn{2}{l!{\color{black!45}\vrule width 0.25pt}}{\textit{SVG-T2I}} & & & & & & & & & \\
Baseline & $\times$1 & 22.30 & 3.83 & \textbf{0.440} & 87.28 & 73.63 & \textbf{0.503} & 51.17 & 23.57 & \textbf{0.475} \\
+ RMG~\cite{curvo2026follow} & $\times$2.5 & 18.73 & 2.17 & 0.429 & 74.31 & 61.71 & 0.501 & 37.52 & 13.80 & \textbf{0.475} \\
\cellcolor{gray!10}+ \method{} (ours) & \cellcolor{gray!10}\textbf{\boldmath$\times$6.25} & \cellcolor{gray!10}\textbf{16.96} & \cellcolor{gray!10}\textbf{1.73} & \cellcolor{gray!10}0.430 & \cellcolor{gray!10}\textbf{50.46} & \cellcolor{gray!10}\textbf{38.80} & \cellcolor{gray!10}0.492 & \cellcolor{gray!10}\textbf{35.09} & \cellcolor{gray!10}\textbf{12.78} & \cellcolor{gray!10}0.468 \\
\midrule
\multicolumn{2}{l!{\color{black!45}\vrule width 0.25pt}}{\textit{SeFi-Image 2B~\cite{feng2026sefi}}} & & & & & & & & & \\
Baseline & $\times$1 & 37.32 & 12.02 & \textbf{0.406} & \textbf{66.64} & \textbf{47.75} & \textbf{0.495} & 79.47 & 43.34 & \textbf{0.440} \\
+ RMG~\cite{curvo2026follow} & $\times$2.5 & 31.70 & \textbf{7.42} & 0.388 & 70.86 & 54.33 & 0.490 & 71.16 & 38.45 & 0.424 \\
\cellcolor{gray!10}+ \method{} (ours) & \cellcolor{gray!10}\textbf{\boldmath$\times$6.25} & \cellcolor{gray!10}\textbf{31.60} & \cellcolor{gray!10}7.70 & \cellcolor{gray!10}0.395 & \cellcolor{gray!10}69.42 & \cellcolor{gray!10}55.09 & \cellcolor{gray!10}0.489 & \cellcolor{gray!10}\textbf{67.66} & \cellcolor{gray!10}\textbf{35.88} & \cellcolor{gray!10}0.435 \\
\bottomrule
\end{tabular}
\end{table*}

\paragraph{Backbones.}
We apply \method to three semantic flow models, each in its released version, with every weight
frozen and nothing adapted to \method. No two share a component:
\textbf{Scale-RAE}~\cite{tong2026scalerae} generates in the latent space of a representation
autoencoder on SigLIP-2~\cite{tschannen2025siglip2}, \textbf{SVG-T2I}~\cite{shi2025svgt2i,shi2025svg}
in its stage-1 release at $256$ pixels, in that of a semantic autoencoder on
DINOv3~\cite{simeoni2025dinov3} and \textbf{SeFi-Image}~\cite{feng2026sefi} in a
DINOv2~\cite{oquab2023dinov2} semantic latent paired with a FLUX.2~\cite{bfl2025flux2} texture
latent, where \method anchors only the semantic latent. Architectures, latent sizes and
checkpoints are in \cref{app:setup}.

\paragraph{Datasets, corpora and retrieval.}
We evaluate on three datasets from different domains: \textbf{COCO}~\cite{lin2014microsoft,chen2015cococaptions}
photographs, \textbf{Anime}~\cite{noneyet2026animecaptions} illustrations and
\textbf{WikiArt}~\cite{atermors2026wikiartrecaption,saleh2015wikiart} paintings. Each is split into a
retrieval corpus and $5{,}000$ held-out caption and image pairs, whose captions are the prompts and
whose images are the FID and KID reference, and \method retrieves from the corpus of the dataset it
is scored on unless stated otherwise. On COCO the corpus is the $118{,}287$ train2017 images, each keyed by the mean SigLIP-2
embedding of its five captions, and the prompts are one caption per val2017 image. Anime and WikiArt
have one caption per image, which is its key. We keep captions that fit SigLIP-2's $64$ tokens and
remove sexual content and duplicates, including corpus near-copies of held-out images, leaving corpora
of $120{,}810$ and $69{,}521$ images respectively (filtering details in \cref{app:setup}).

\paragraph{Metrics.}
Fidelity is FID~\cite{heusel2017gans} and KID~\cite{binkowski2018demystifying}, its unbiased
counterpart on the same Inception features, against the $5{,}000$ held-out images of the
dataset. Prompt alignment is the cosine similarity of OpenCLIP
ViT-bigG-14~\cite{cherti2023reproducible,radford2021learning} between image and caption. Precision and
recall~\cite{kynkaanniemi2019improved} and density and coverage~\cite{naeem2020reliable} are in \cref{tab:app_prdc}.
Per-prompt diversity, as in a diversity-oriented
noise optimisation~\cite{shi2026mono}, is scored on four samples for each of the first $1{,}000$
COCO prompts by their mean pairwise LPIPS~\cite{zhang2018unreasonable} and DreamSim~\cite{fu2023dreamsim}
and their Vendi score~\cite{friedman2023vendi}, the effective number of distinct images among the
four, higher is more diverse. Because the anchor is built from real images, we also audit
memorisation~\cite{carlini2023extracting,somepalli2023diffusion} by the near-duplicate rate, the share
of generations whose SSCD similarity~\cite{pizzi2022self} to their nearest corpus image exceeds $0.5$,
the threshold of \citet{somepalli2023diffusion}, against the same rate for the real held-out images.

\paragraph{The baselines and the operating point.}
Unless a table states otherwise, our baseline is each backbone's released sampler, $50$ steps from
Gaussian noise at its default guidance scale, $w = 1$ for Scale-RAE and $w = 4$ for SVG-T2I and
SeFi-Image, with the same prompts and the same draw of $\veps$ per prompt as \method.
We found no training-free method that steers sampling with reference images in a semantic
latent space, so we compare with the closest one in a VAE space, Reference-Mean Guidance
(RMG)~\cite{curvo2026follow}, applied unchanged in the semantic spaces of our backbones in its
published setting, $20$ steps at $w = 3.5$ (from their code), and we use it with the top $20$ retrieved images as its bank. Its one
free parameter, the strength $\gamma_0$, we set per backbone and dataset to the one with the lowest KID of the three
values we test (see \cref{app:setup}). Unless stated otherwise \method uses $k = 50$,
$N = 8$ steps and $w = 4$, with $\tau = 0.02$ and $t_0 = 0.05$.
We report speedup in sampling steps. A guided step ($w > 1$) takes two network evaluations, which run
in parallel, so at the same $w$ as the released sampler \method costs the same per step. Only
Scale-RAE's released sampler is unguided, so there \method is $3.125\times$ faster in network
evaluations, not $6.25\times$.

\subsection{Main results}
\label{sec:exp:fewstep}
\Cref{tab:main} gives the main result: without training anything, and with $8$ steps against the
released sampler's $50$, \method reaches a lower FID and KID on every backbone and dataset but one,
SeFi-Image on Anime, where the retrieved latents are almost as mixed in with the rest of the corpus as in a VAE space,
so their mean says less about the prompt (\cref{sec:exp:control}).
Prompt alignment stays close to the released sampler's, slightly higher on Scale-RAE and slightly
lower on SVG-T2I and SeFi-Image.
RMG takes $2.5\times$ the steps, but \method still reaches a lower FID and KID than RMG on every dataset for
Scale-RAE and SVG-T2I, and on WikiArt for SeFi-Image. On SeFi-Image the two are level on COCO, and
\method is ahead in FID on Anime, where neither reaches the released sampler. Alignment does not
separate the two consistently.

Several checks, detailed in \cref{app:results}, show that the gain is not due to the
setup. We tested different step counts on COCO, and at all equal step counts \method
has a lower FID than the released sampler on all three backbones. We also test the DPM-Solver++
solver~\cite{lu2022dpmsolverpp} instead of Euler, and the gain is almost unchanged.
Furthermore, recall and coverage on COCO, which fall when a sampler drops modes, stay level or
rise on all three backbones. We also checked if the gain is from copying and find that the near-duplicate rate of \method stays
within a fraction of a point of the released sampler's, far below that of the real held-out images.
To check the gain beyond the Inception features of FID, we
use FD$_\text{DINOv2}$~\cite{stein2023exposing} and CMMD~\cite{jayasumana2024rethinking}, which compare
the same images in the semantic feature spaces of DINOv2 and CLIP. \method has the lowest FD$_\text{DINOv2}$ in
all nine settings and the lowest CMMD in eight of them. \Cref{app:fd_cmmd} describes both
metrics and gives their values.
Finally, we test RMG and \method at the same $k$, and their comparison is unchanged.

\subsection{Diversity}
\label{sec:exp:diversity}

\begin{figure}[t]
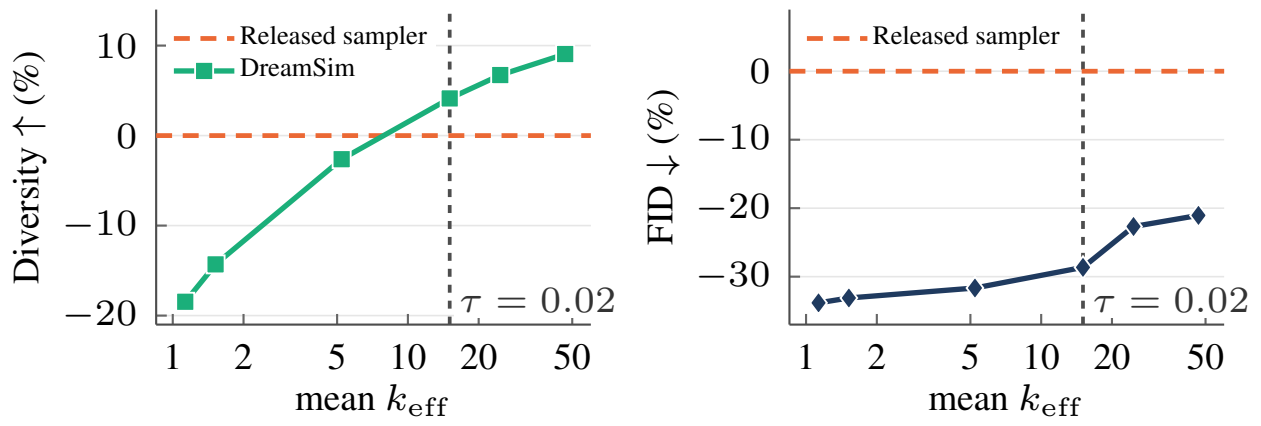

  \centering
  \input{figures/div_keff/values}%
  \input{figures/div_keff/style}%
  \ifdefined\figcachebegin\figcachebegin{div_keff}\fi
  \dkpanel{rae}{dkdiv}{ylabel={Diversity $\uparrow$ (\%)}, ymax=14, dkleginside,
    legend entries={Released sampler, DreamSim}}{\dkplot{rae}{dreamsim}}{}{$\tau = 0.02$}\hfill
  \dkpanel{rae}{dkqual}{ylabel={FID $\downarrow$ (\%)}, ymax=9, dkleginside,
    legend entries={Released sampler}}{\dkplot{rae}{fid}}{}{$\tau = 0.02$}
  \ifdefined\figcacheend\figcacheend\fi
  \caption{\textbf{Diversity and FID against the effective number of neighbours.} \method on Scale-RAE as
    $\tau$ varies at $k = 50$, against $k_\mathrm{eff}$ (\cref{sec:exp:diversity}): DreamSim diversity
    (left) and FID-5k (right), each as the change from the released
    sampler (dashed). More effective neighbours means higher diversity but costs
    part of the FID gain. LPIPS and Vendi follow the same pattern. The dotted line marks our $\tau = 0.02$, where diversity is back at the released
    sampler's level.
    }
  \label{fig:div_keff}
\end{figure}

An anchor built from real images could make a prompt produce nearly the same image every time.
Mode-seeking objectives~\cite{thamizharasan2025meanshift} aim at a single mode and may give up variety,
so we test whether the anchor does so inadvertently.
On COCO, \method is more diverse than the released sampler on all three backbones by DreamSim ($+4$
to $+18\%$) and Vendi ($+3$ to $+15\%$), and within $2\%$ by LPIPS. It is also more diverse than RMG on
every measure on Scale-RAE and SVG-T2I, and about level with it on SeFi-Image, while on Scale-RAE RMG
is even less diverse than the released sampler. \Cref{app:diversity} reports the full numbers
and shows several samples per prompt from each method.

How much variety the anchor leaves depends on how many neighbours it averages. Because the mean is
weighted, that number can be much smaller than $k$, and when $\tau \to 0$ only the nearest neighbour
gets all the weight. We measure it with
Kish's effective sample size~\cite{kish1965survey} of the weights $\alpha_j$ in \cref{eq:anchor},
$k_\mathrm{eff} = 1 / \sum_j \alpha_j^2$, which equals $k$ when the weights are uniform and $1$ when a
single neighbour takes all the weight. Both $\tau$ and $k$ change it, but in
our sweeps $k$ changes the results little once $k > 5$ (see \cref{tab:ablation}), so we use
$\tau$ as the knob. \Cref{fig:div_keff} shows that on Scale-RAE, $k_\mathrm{eff}$ trades diversity against
FID. A concentrated anchor gives the lowest FID but less diversity than the released sampler, which we
attribute to the samples of a prompt being drawn around fewer real images. Spreading the weight over
more neighbours raises diversity, and the FID gain shrinks. The $\tau$ sweep in
\cref{fig:keff_backbones} shows the same trend on SVG-T2I, and a much weaker one on SeFi-Image, where
FID and diversity barely change with $\tau$, possibly because the anchor enters only its semantic latent. We set $\tau = 0.02$, the most
concentrated anchor we tested at which Scale-RAE's diversity is on par with the released sampler's or
above it, and which keeps most of the FID gain.

\subsection{Why use a semantic space?}
\label{sec:exp:control}

\begin{figure}[t]
  \centering
  \input{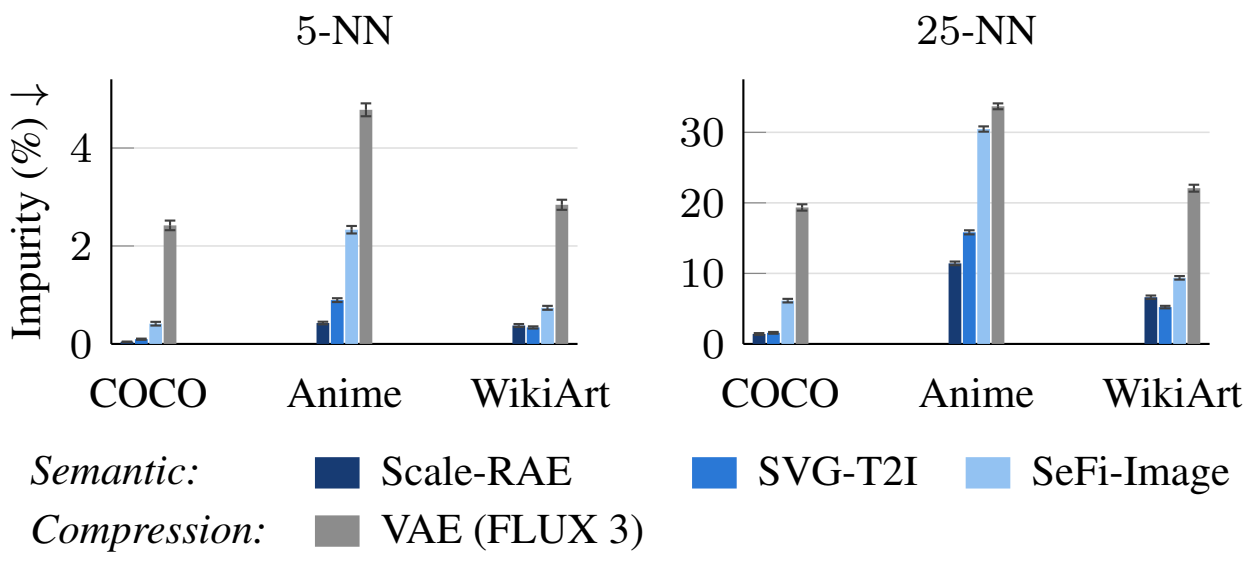}%
  \newcommand{\imw}{0.37\linewidth}%
  \newcommand{\imh}{0.21\linewidth}%
  \newcommand{\imbar}{2.4pt}%
  \newcommand{\imgap}{0.3pt}%
  \newcommand{\imfont}{\footnotesize}%
  \newcommand{\imlegfont}{\footnotesize}%
  \definecolor{imrae}{HTML}{173B73}%
  \definecolor{imsvg}{HTML}{2A78D6}%
  \definecolor{imsefi}{HTML}{93C2F2}%
  \definecolor{imvae}{HTML}{8C8C8C}%
  \newcommand{\imv}[3]{\csname imp@#1@#2@#3\endcsname}%
  \pgfplotsset{imaxis/.style={ybar, bar width=\imbar, height=\imh, scale only axis, clip=false, ymin=0,
      axis x line*=bottom, axis y line*=left, ymajorgrids, grid style={black!12, line width=0.3pt},
      tick label style={font=\imfont}, label style={font=\imfont}, title style={font=\imfont, yshift=-3pt},
      ylabel style={yshift=-3pt},
      x tick label style={anchor=north, inner sep=2pt},%
      major x tick style={draw=none}, scaled y ticks=false,
      yticklabel style={/pgf/number format/fixed, /pgf/number format/precision=1}},
    imerr/.style={error bars/.cd, y dir=both, y explicit,
      error bar style={line width=0.4pt, black!75},
      error mark options={rotate=90, mark size=1pt, line width=0.4pt, black!75}},
    imswatch/.style={legend image code/.code={\fill[#1] (0pt,-2.5pt) rectangle (8pt,3pt);}}}%
  \newcommand{\imbars}[2]{%
    \foreach \s/\c/\b in {scale_rae/imrae/-1.5, svg_t2i/imsvg/-0.5, sefi/imsefi/0.5, flux256/imvae/1.5} {
      \def\imcoords{}%
      \foreach \lab/\ds in {#2} {\xdef\imcoords{\imcoords (\lab,\imv{\s}{\ds}{#1}) +- (0,\imv{\s}{\ds}{#1@ci})}}%
      \edef\implot{\noexpand\addplot[fill=\c, draw=none, bar shift=\b*(\imbar+\imgap), imerr]
        coordinates {\imcoords};}\implot}}%
  \begin{tikzpicture}[baseline=(imax.south)]
    \begin{axis}[imaxis, name=imax, width=\imw, title={$5$-NN}, ylabel={Impurity (\%) $\downarrow$},
        symbolic x coords={COCO,Anime,WikiArt}, xtick=data, enlarge x limits={abs=7pt},
        legend to name=imlegend, legend columns=4,%
        legend cell align=left,
        legend style={draw=none, font=\imlegfont, /tikz/every odd column/.append style={column sep=2pt},
          /tikz/every even column/.append style={column sep=7pt}}]
      \addlegendimage{empty legend}\addlegendentry{\textit{Semantic:}}
      \addlegendimage{imswatch=imrae}\addlegendentry{Scale-RAE}
      \addlegendimage{imswatch=imsvg}\addlegendentry{SVG-T2I}
      \addlegendimage{imswatch=imsefi}\addlegendentry{SeFi-Image}
      \addlegendimage{empty legend}\addlegendentry{\textit{Compression:}}
      \addlegendimage{imswatch=imvae}\addlegendentry{VAE (FLUX 3)}
      \addlegendimage{empty legend}\addlegendentry{}
      \addlegendimage{empty legend}\addlegendentry{}
      \imbars{5}{COCO/coco,Anime/anime,WikiArt/art}
    \end{axis}
  \end{tikzpicture}\hfill
  \begin{tikzpicture}[baseline=(imax.south)]
    \begin{axis}[imaxis, name=imax, width=\imw, title={$25$-NN},
        symbolic x coords={COCO,Anime,WikiArt}, xtick=data, enlarge x limits={abs=7pt}]
      \imbars{25}{COCO/coco,Anime/anime,WikiArt/art}
    \end{axis}
  \end{tikzpicture}
  \\[2pt]
  \pgfplotslegendfromname{imlegend}
  \caption{\textbf{Impurity of the retrieved neighbours}. The share of the corpus
    closer to a prompt's anchor than its $n$-th retrieved neighbour, in the semantic latent spaces and
    the compression VAE latent space. Mean
    over each dataset's $5{,}000$ held-out prompts.}
  \label{fig:impurity}
\end{figure}

\begin{table}[t]
  \centering
  \small
  \caption{\textbf{The anchor as a replacement for guidance.} With guidance is the setting of \cref{tab:main}, without guidance is $w = 0$. Speedup counts sampler steps, as in \cref{tab:main}. At $w = 0$ each step also takes one network evaluation instead of the two that guidance needs. Best per column at each guidance setting in bold.}
  \label{tab:uncond}
  \setlength{\tabcolsep}{2.5pt}
  \footnotesize
  \begin{tabular}{llccccc}
    \toprule
    & & & \multicolumn{2}{c}{Scale-RAE} & \multicolumn{2}{c}{SVG-T2I} \\
    \cmidrule(lr){4-5} \cmidrule(lr){6-7}
    & & Speedup $\uparrow$ & FID $\downarrow$ & CLIP $\uparrow$ & FID $\downarrow$ & CLIP $\uparrow$ \\
    \midrule
     & Baseline & $\times$1 & 29.51 & 0.418 & 22.30 & \textbf{0.440} \\
    \multirow{-2}{*}{W/ guidance} & \cellcolor{gray!10}+ \method & \cellcolor{gray!10}\textbf{$\times$6.25} & \cellcolor{gray!10}\textbf{21.06} & \cellcolor{gray!10}\textbf{0.420} & \cellcolor{gray!10}\textbf{16.96} & \cellcolor{gray!10}0.430 \\
    \midrule
     & Baseline & $\times$1 & 63.14 & 0.112 & 60.02 & 0.103 \\
    \multirow{-2}{*}{W/o guidance} & \cellcolor{gray!10}+ \method & \cellcolor{gray!10}\textbf{$\times$6.25} & \cellcolor{gray!10}\textbf{27.97} & \cellcolor{gray!10}\textbf{0.345} & \cellcolor{gray!10}\textbf{17.72} & \cellcolor{gray!10}\textbf{0.344} \\
    \bottomrule
  \end{tabular}
  \small
  \par\vspace{8pt}%
{\centering
  \newcommand{\gpanelw}{0.24}%
  \newcommand{\gcolsep}{0.8pt}%
  \newcommand{\gpairsep}{4pt}%
  \newcommand{\glabsize}{\footnotesize}%
  \newcommand{\gheadsize}{\footnotesize}%
  \newcommand{\gheadfont}{\usefont{T1}{ppl}{m}{it}}%
  \newcommand{\gheadsep}{1pt}%
  \newcommand{\glabsep}{1pt}%
  \definecolor{uqbase}{HTML}{EB6834}%
  \definecolor{uqours}{HTML}{2A78D6}%
  \setlength{\tabcolsep}{\gcolsep}%
  \renewcommand{\arraystretch}{0}%
  \newcommand{\gtile}[2]{\includegraphics[width=\gpanelw\linewidth]{figures/uncond_qual/#1/#2.png}}%
  \newcommand{\ghead}[1]{\multicolumn{2}{c}{\gheadsize\gheadfont\strut #1}}%
  \newcommand{\glab}[2]{\hbox to \gpanelw\linewidth{\hss\glabsize\color{#1}\strut #2\hss}}%
  \begin{tabular}{@{}cc@{\hspace{\gpairsep}}cc@{}}
    \ghead{W/o guidance} & \ghead{W/ guidance} \\
    \gtile{p01288}{base_wo} & \gtile{p01288}{rsd_wo} & \gtile{p01288}{base_w} & \gtile{p01288}{rsd_w} \\
    \noalign{\vskip\glabsep}
    \glab{uqbase}{Baseline} & \glab{uqours}{\textbf{\method}} & \glab{uqbase}{Baseline} & \glab{uqours}{\textbf{\method}} \\
  \end{tabular}\par\vspace{2pt}%
  
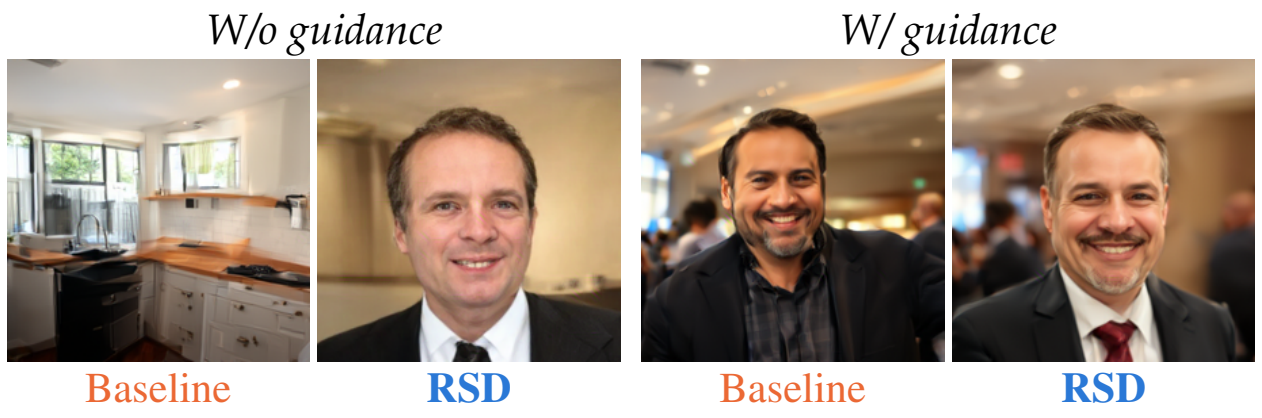
\captionof{figure}{\textbf{The anchor without guidance.} The four cells of \cref{tab:uncond} for
  \gridprompt{A man in a suit smiling at the camera} on Scale-RAE. At $w = 0$ the prompt reaches the
  sampler only through the anchor of \method, which is enough to draw a man in a suit.}
  \label{fig:uncond_qual}\par}

\end{table}

\method assumes that images with close captions also have close latents, so that the mean of
the retrieved latents aligns with the prompt.
This is less likely to hold for models that generate in the latent space of an autoencoder trained for
pixel reconstruction, which places images by appearance rather than content.
We therefore measure how tightly the retrieved latents group in each kind of space.

We define the
\textbf{$n$-NN impurity} of a prompt as the share of the corpus that lies closer to its anchor $\vmu$ than the
$n$-th nearest of its retrieved latents, by L2 distance in the backbone's latent space.
A high impurity means the retrieved latents are scattered among unrelated corpus latents, so their mean is
no closer to the prompt's content than much of the rest of the corpus. A low impurity
means they form a tight group whose mean stands for their common content. Our settings
average over about $5$ to $25$ effective neighbours (\cref{fig:div_keff}), so we test $n \in \{5, 25\}$, comparing the three semantic
spaces with the latent space of the FLUX 3 autoencoder~\cite{bfl2026flux3klein}, similar to the space RMG was introduced in. On every
dataset and at every $n$ the VAE space is the most impure (\cref{fig:impurity}), $5$ to $56$ times more
than Scale-RAE and SVG-T2I at $n = 5$ and $2$ to $14$ times at $n = 25$.
So in a non-semantic space a mean of neighbours is a weaker description of the prompt.
On the one setting where \method falls behind the released sampler in FID, SeFi-Image on Anime,
this semantic space impurity is the closest to the VAE space, almost level
with it at $n = 25$, which suggests that there, the anchor describes the prompt poorly.

\subsection{Conditioning without guidance}
\label{sec:exp:cfgfree}
The anchor can also stand in for the prompt. With $w = 0$ in \cref{eq:cfg} the sampler uses only the
unconditional velocity, so the prompt reaches the image only through the initial state
(\cref{tab:uncond,fig:uncond_qual}). We test this on Scale-RAE and SVG-T2I, since SeFi-Image's
unconditional branch alone gives no usable images (FID $168$). From Gaussian noise, both samplers
ignore the prompt, with CLIP near $0.1$. Started from the anchor, they close about three quarters of
the CLIP gap to the released sampler, and their FID is already lower than the released
sampler's, which does see the prompt.

\subsection{Ablations}
\label{sec:exp:ablations}

\begin{table}[t]
\centering
\footnotesize
\setlength{\tabcolsep}{2.5pt}
\renewcommand{\arraystretch}{1.1}
\caption{\textbf{Corpus against held-out set.} FID-5k and CLIP of \method on each dataset's held-out prompts (blocks) with anchors from each corpus (rows), the dataset's own corpus shaded.}
\label{tab:cross}
\begin{tabular}{lcccccc}
\toprule
 & \multicolumn{2}{c}{Scale-RAE} & \multicolumn{2}{c}{SVG-T2I} & \multicolumn{2}{c}{SeFi-Image} \\
\cmidrule(lr){2-3} \cmidrule(lr){4-5} \cmidrule(lr){6-7}
Corpus & FID$\downarrow$ & CLIP$\uparrow$ & FID$\downarrow$ & CLIP$\uparrow$ & FID$\downarrow$ & CLIP$\uparrow$ \\
\midrule
\multicolumn{7}{c}{\textit{COCO held-out set (prompts)}} \\
\cellcolor{gray!10}COCO & \cellcolor{gray!10}\textbf{21.06} & \cellcolor{gray!10}\textbf{0.420} & \cellcolor{gray!10}\textbf{16.96} & \cellcolor{gray!10}\textbf{0.430} & \cellcolor{gray!10}\textbf{31.60} & \cellcolor{gray!10}\textbf{0.395} \\
Anime & 68.98 & 0.358 & 64.64 & 0.363 & 40.84 & 0.373 \\
WikiArt & 61.89 & 0.358 & 42.56 & 0.384 & 34.29 & 0.386 \\
\midrule
\multicolumn{7}{c}{\textit{Anime held-out set (prompts)}} \\
COCO & 88.48 & 0.430 & 100.82 & 0.475 & 76.58 & 0.484 \\
\cellcolor{gray!10}Anime & \cellcolor{gray!10}\textbf{49.84} & \cellcolor{gray!10}\textbf{0.467} & \cellcolor{gray!10}\textbf{50.46} & \cellcolor{gray!10}\textbf{0.492} & \cellcolor{gray!10}\textbf{69.42} & \cellcolor{gray!10}\textbf{0.489} \\
WikiArt & 75.27 & 0.459 & 78.53 & 0.489 & 72.29 & 0.487 \\
\midrule
\multicolumn{7}{c}{\textit{WikiArt held-out set (prompts)}} \\
COCO & 57.91 & 0.410 & 56.97 & 0.433 & 67.76 & 0.426 \\
Anime & 67.41 & 0.428 & 58.16 & 0.438 & 69.55 & 0.431 \\
\cellcolor{gray!10}WikiArt & \cellcolor{gray!10}\textbf{32.07} & \cellcolor{gray!10}\textbf{0.458} & \cellcolor{gray!10}\textbf{35.09} & \cellcolor{gray!10}\textbf{0.468} & \cellcolor{gray!10}\textbf{67.66} & \cellcolor{gray!10}\textbf{0.435} \\
\bottomrule
\end{tabular}
\end{table}

\paragraph{The anchor.}
\label{sec:exp:anchor}
Since the corpus, the prompts and the FID reference all come from the same dataset, one may ask
whether any start typical of that dataset would lower FID, whatever the prompt. We show that it does not. When each
prompt keeps its text condition but gets the anchor retrieved for another, randomly drawn prompt, FID
is worse than the released sampler's, $38.63$ against $29.51$ on Scale-RAE and $24.66$ against
$22.30$ on SVG-T2I, and alignment drops, CLIP $0.274$ and $0.293$ against $0.418$ and $0.440$. The corpus mean and a random
single corpus image as anchor are also worse than the released sampler (see \cref{tab:anchor_app}).
The corpus helps the most when its images resemble what the prompts describe: of the three corpora,
the one from the prompt's domain gives the best FID and CLIP on every backbone and dataset
(\cref{tab:cross}).
This also bounds our claim. \method does not make images semantically better in general, but aligns them
with the semantic domain of the corpus, which helps the prompts from that domain.

\begin{figure}[t]
  \centering
  \input{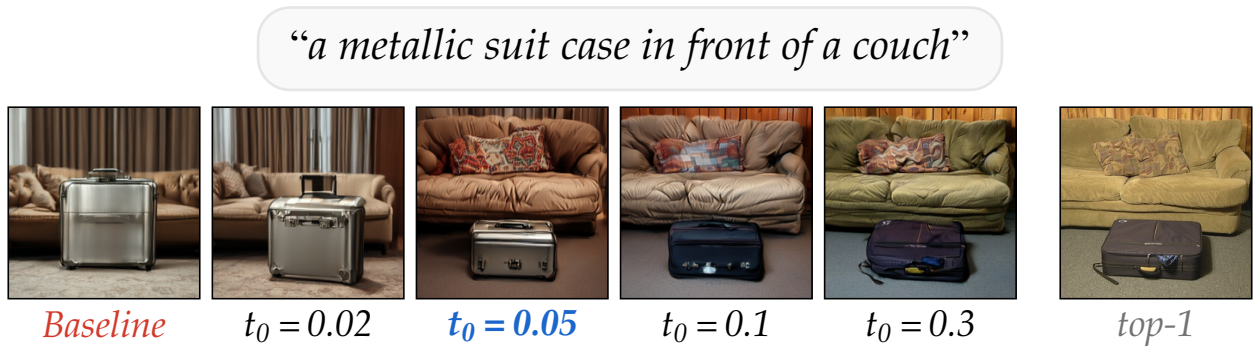}%
  \newcommand{\tzside}{0.152\linewidth}%
  \newcommand{\tzgap}{0.01\linewidth}%
  \newcommand{\tznbgap}{0.035\linewidth}%
  \newcommand{\tzlabsep}{2pt}%
  \newcommand{\tzpromptsep}{3pt}%
  \newcommand{\tzpromptfont}{\footnotesize}%
  \newcommand{\tzpromptfill}{black!2.5}%
  \newcommand{\tzframe}{0.3pt}%
  \newcommand{\tzfont}{\scriptsize\usefont{T1}{ppl}{m}{it}}%
  \newcommand{\tzbasefont}{\scriptsize\usefont{T1}{ppl}{m}{it}}%
  \newcommand{\tzoursfont}{\scriptsize\usefont{T1}{ppl}{b}{it}}%
  \definecolor{tzbase}{HTML}{D33A2C}%
  \newcommand{\tzimg}[1]{\includegraphics[width=\tzside,height=\tzside]{figures/t0_strip/#1}}%
  \newcommand{\tzlab}[1]{t\textsubscript{0}\,=\,#1}%
  \ifdefined\figcachebegin\figcachebegin{t0_strip}\fi
  \begin{tikzpicture}[panel/.style={inner sep=0pt, outer sep=0pt},
                      lab/.style={inner sep=0pt, anchor=north}]
    \foreach \t [count=\i] in \tzslist {
      \ifnum\i=1
        \node[panel, anchor=south west] (c\i) at (0,0) {\tzimg{t0_\t.png}};
      \else
        \node[panel, anchor=west] (c\i) at ($(c\the\numexpr\i-1\relax.east)+(\tzgap,0)$) {\tzimg{t0_\t.png}};
      \fi
      \draw[black, line width=\tzframe] (c\i.south west) rectangle (c\i.north east);
      \edef\tzcur{\t}%
      \ifnum\i=1
        \node[lab, font=\tzbasefont] at ($(c\i.south)+(0,-\tzlabsep)$)
          {\textcolor{tzbase}{Baseline}};
      \else\ifx\tzcur\tzsours
        \node[lab, font=\tzoursfont, text=otherblue] at ($(c\i.south)+(0,-\tzlabsep)$) {\tzlab{\t}};
      \else
        \node[lab, font=\tzfont] at ($(c\i.south)+(0,-\tzlabsep)$) {\tzlab{\t}};
      \fi\fi
      \xdef\tzlast{c\i}%
    }
    \node[panel, anchor=west] (nb) at ($(\tzlast.east)+(\tznbgap,0)$) {\tzimg{top1.jpg}};
    \draw[black, line width=\tzframe] (nb.south west) rectangle (nb.north east);
    \node[lab, font=\tzfont, text=black!55] at ($(nb.south)+(0,-\tzlabsep)$) {top-1};
    \node[anchor=south, fill=\tzpromptfill, draw=black!10, line width=0.5pt, rounded corners=6pt, inner xsep=6pt,
          inner ysep=3pt, align=center, font=\tzpromptfont] at ($(c1.north west)!0.5!(nb.north east)+(0,\tzpromptsep)$)
      {\gridprompt{\tzsprompt}};
  \end{tikzpicture}%
  \ifdefined\figcacheend\figcacheend\fi
  \vspace{-4pt}%
  \caption{\textbf{How much noise to add.} \method's sample on Scale-RAE as $t_0$ grows, from Gaussian noise
    (left, $t_0 = 0$ with $50$ steps) towards the top-1 retrieved image (right). We use $t_0 = \tzsours$.}
  \label{fig:t0_strip}
  \vspace{-6pt}%
\end{figure}

\paragraph{The displacement $t_0$.}
\label{sec:exp:t0}
As \cref{fig:t0_strip} shows, the displacement $t_0$ of \cref{eq:state} sets how far the start moves
from Gaussian noise towards the anchor, and with it how close the sample comes to the retrieved images.
Up to $t_0 = 0.05$ the suitcase stays metallic, as the prompt asks, and from $t_0 = 0.1$ it takes the dark
fabric of the retrieved image. The sweep in \cref{tab:t0_knob} shows alignment falls
steadily as $t_0$ grows, diversity peaks at $t_0 = 0.05$, and from $t_0 = 0.1$ the near-duplicate rate rises
above the released sampler's, so we use $t_0 = 0.05$. $t_0$ can also be
set per use, for example on SVG-T2I, a larger $t_0$ trades alignment for a lower FID,
and a smaller one keeps the
released sampler's alignment while still lowering its FID but trades diversity.

\begin{figure}[t]
  \centering
  \input{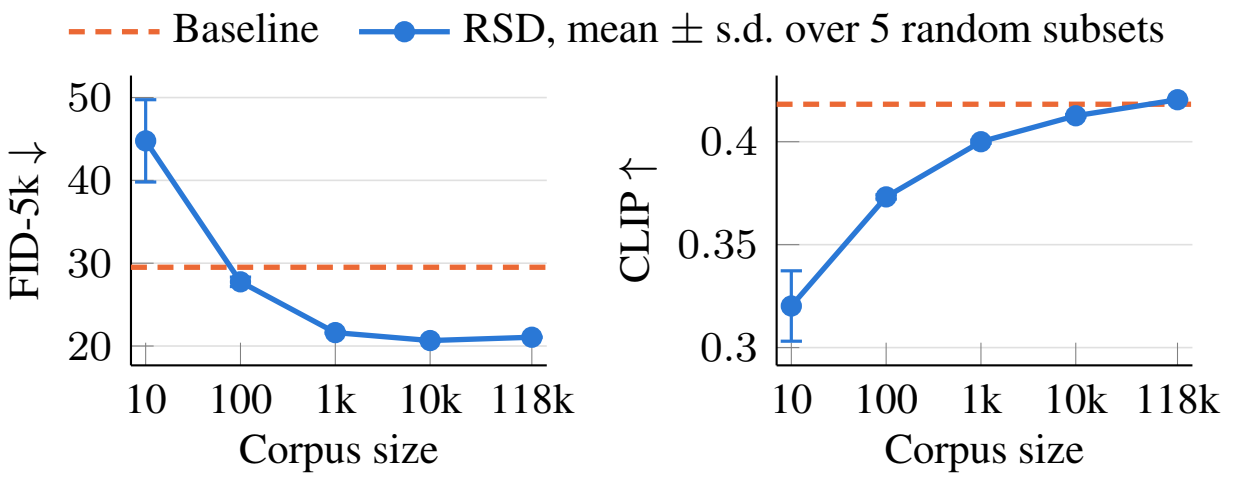}%
  \newcommand{\csw}{0.33\linewidth}%
  \newcommand{\csh}{0.23\linewidth}%
  \newcommand{\csfont}{\footnotesize}%
  \newcommand{\cslegfont}{\footnotesize}%
  \definecolor{csours}{HTML}{2A78D6}%
  \definecolor{csbase}{HTML}{EB6834}%
  \newcommand{\csv}[1]{\csname csz@#1\endcsname}%
  \pgfplotsset{csaxis/.style={width=\csw, height=\csh, scale only axis,
      xmin=7, xmax=1.7e5, xtick={10,100,1000,10000,118287}, xticklabels={10,100,1k,10k,118k},
      minor xtick={}, axis x line*=bottom, axis y line*=left, ymajorgrids, grid style={black!12, line width=0.3pt},
      tick label style={font=\csfont}, label style={font=\csfont}, xlabel={Corpus size}, xlabel style={yshift=3pt}, ylabel style={yshift=-3pt},
      scaled y ticks=false, yticklabel style={/pgf/number format/fixed}},
    csm/.style={csours, line width=1pt, mark=*, mark size=1.5pt,%
      error bars/.cd, y dir=both, y explicit, error bar style={line width=0.6pt, csours},
      error mark options={rotate=90, mark size=2pt, line width=0.6pt, csours}},
    csb/.style={csbase, line width=1pt, dash pattern=on 3pt off 2pt}}%
  \newcommand{\cspanel}[3]{%
    \begin{tikzpicture}[baseline=(csax.south)]
      \begin{axis}[xmode=log, csaxis, name=csax,%
          ylabel={#2}, yticklabel style={/pgf/number format/precision=#3},
          legend to name=cslegend#1, legend columns=2,
          legend style={draw=none, font=\cslegfont, /tikz/every even column/.append style={column sep=5pt}}]
        \edef\csplot{\noexpand\addplot[csb, domain=7:1.7e5, samples=2] {\csv{#1@base}};
                     \noexpand\addlegendentry{Baseline}
                     \noexpand\addplot[csm] coordinates {\csv{#1@mean}};
                     \noexpand\addlegendentry{\noexpand\method{}, mean $\pm$ s.d.\ over \csv{ndraws} random subsets}}\csplot
      \end{axis}
    \end{tikzpicture}}%
  \pgfplotslegendfromname{cslegendfid}\\[-2pt]
  \cspanel{fid}{FID-5k $\downarrow$}{0}\hfill\cspanel{clip}{CLIP $\uparrow$}{2}
  \vspace{-4pt}%
  \caption{\textbf{Corpus size.} FID-5k and CLIP of \method on Scale-RAE retrieving from a random subset of the
    COCO corpus. The last point is the full
    corpus and the dashed line the released sampler.}
  \label{fig:corpus_size}
\end{figure}

\paragraph{Corpus size.}
\label{sec:exp:corpus_size}
We also test how large the corpus must be. On Scale-RAE we retrieve from random subsets of the COCO corpus, five
subsets per size (\cref{fig:corpus_size}). $100$ images already beat the released sampler's FID
($27.8 \pm 0.6$ against $29.5$), while alignment keeps rising with the size of the corpus and reaches the released sampler
only at the full corpus. The subsets of each size
agree closely from $100$ images on, and no generation copies an image of its subset (SSCD above $0.5$).

\section{Conclusions}
\label{sec:conclusion}

Text-to-image flow models start every sample from the same Gaussian, whatever the prompt. We asked
whether a corpus of captioned images from a target domain can turn it into a prompt-dependent prior
that steers a frozen semantic flow model towards that domain, and found that it can. \method shifts
the initial noise a short step towards a similarity-weighted mean of the latents of the corpus images
retrieved for the prompt, and trains nothing. With $6.25\times$ fewer sampling steps, it lowers FID
to the target dataset in eight of nine settings with comparable prompt alignment, and it matches or
beats the FID of RMG in $2.5\times$ fewer steps. Our analysis suggests why it works: in a semantic
latent space the retrieved neighbours have a low impurity, so their mean still describes the prompt,
and a corpus drawn from the target domain gives the largest gain. Since nothing is trained, changing
the backbone or the domain only means encoding a new corpus, and every sample comes with the corpus
images that seeded it. We discuss the limitations of \method in \cref{app:limitations}.

{
    \small
    \bibliographystyle{ieeenat_fullname}
    \bibliography{main}
}

\input{sections/suppl}

\end{document}